%% file: arxiv.tex
\documentclass{article}
\usepackage{iclr2026/iclr2026_conference,times}
\iclrfinalcopy
\usepackage{hyperref}
\usepackage{url}

\usepackage{colortbl}
\usepackage{threeparttable}

\makeatletter
  \newcommand\figcaption{\def\@captype{figure}\caption}
  \newcommand\tabcaption{\def\@captype{table}\caption}
\makeatother

\usepackage[utf8]{inputenc} 
\usepackage[T1]{fontenc}    
\usepackage{url}            
\usepackage{booktabs}       
\usepackage{amsfonts}       
\usepackage{nicefrac}       
\usepackage{microtype}      
\usepackage{xcolor}         
\usepackage{bbding}         
\usepackage{graphicx}
\usepackage{tabularx}       
\usepackage{comment}
\usepackage{subcaption}
\usepackage{caption}

\usepackage{amsmath, amsfonts}
\usepackage{amssymb}
\usepackage{float}
\usepackage{multirow}

\usepackage[capitalize]{cleveref}
\crefname{section}{Sec.}{Secs.}
\Crefname{section}{Section}{Sections}
\Crefname{table}{Table}{Tables}
\crefname{table}{Tab.}{Tabs.}

\input{comments_tool/macros.tex}
\title{\fontsize{16pt}{9pt}\selectfont MotionGPT3: Human Motion as a Second Modality}
\author{
    Bingfan Zhu$^{1}$\quad
    Biao Jiang$^{2}$\quad
    Sunyi Wang$^{1}$\quad
    Shixiang Tang$^{4}$\quad
    Tao Chen$^{2}$\quad
    \\
    \textbf{Linjie Luo}$^{3}$\quad
    \textbf{Youyi Zheng}$^{1}$\textsuperscript{\dag} \quad
    \textbf{Xin Chen}$^{3}$\textsuperscript{\dag}
    \\
    $^{1}$Zhejiang University \quad
    $^{2}$Fudan University \quad
    $^{3}$ByteDance \quad
    \\
    $^{4}$The Chinese University of HongKong \\
    {\tt \small \textbf{\href{https://github.com/OpenMotionLab/MotionGPT3}{https://github.com/OpenMotionLab/MotionGPT3}}} \\
}

\begin{document}

\maketitle


\input{sections/abstract.tex}
\input{sections/introduction.tex}

\input{sections/related.tex}

\input{sections/method.tex}
\input{sections/experiment.tex}

\input{sections/discussion}

\newpage
{
\small
\bibliography{arxiv}
\bibliographystyle{iclr2026/iclr2026_conference}
}

\newpage
\vspace{-4pt}
\section*{\Large Appendix}
\input{sections/appendix.tex}

\input{sections/LLM.tex}




\end{document}

%% file: comments_tool/macros.tex
\usepackage{hyphenat}
\usepackage{soul,color}
\usepackage{amsopn}

\definecolor{Red}{cmyk}{0,1,1,0}
\definecolor{Green}{cmyk}{1,0,1,0}
\definecolor{Cyan}{cmyk}{1,0,0,0}
\definecolor{Purple}{cmyk}{0.45,0.86,0,0}
\definecolor{Rosolic}{cmyk}{0.00,1.00,0.50,0}
\definecolor{Blue}{cmyk}{1.00,1.00,0.00,0}
\definecolor{BlueViolet}{cmyk}{0.86,0.91,0,0.04}
\definecolor{NavyBlue}{cmyk}{0.94,0.54,0,0}
\definecolor{Violet}{cmyk}{0.79,0.88,0,0}
\definecolor{RoyalPurple}{cmyk}{0.75,0.90,0,0}
\definecolor{MidnightBlue}{cmyk}{0.98,0.13,0,0.43}
\definecolor{SkyBlue}{cmyk}{0.62,0,0.12,0}
\definecolor{BlueGreen}{cmyk}{0.85,0,0.33,0}
\definecolor{SeaGreen}{cmyk}{0.69,0,0.50,0}

\definecolor{textC}{cmyk}{0.94,0.54,0,0} 
\definecolor{motionC}{cmyk}{0,0.51,1,0}

\newcommand{\Ours}{MotionGPT3}

\newcommand{\hidden}[1]{{\color{NavyBlue}}}

%% file: sections/abstract.tex

\vspace{-6pt}
\begin{abstract}
\vspace{-6pt}
With the rapid progress of large language models (LLMs), multimodal frameworks that unify understanding and generation have become promising, yet they face increasing complexity as the number of modalities and tasks grows. 
We observe that motion quantization introduces approximation errors that cap motion quality, while unifying discrete text and continuous motion within a single-stream backbone amplifies cross-modal interference. 
Motivated by recent multi-branch designs that separate signals from different modalities, we propose \Ours, a bimodal motion–language model for both understanding and generation. 
\Ours~encodes raw motion into a continuous latent space, thereby avoiding quantization-induced artifacts, while leveraging the semantic prior of pretrained language models.
A dual-stream Transformer with shared attention preserves modality-specific routes while enabling controlled, bidirectional information flow, which reduces interference, stabilizing optimization, 
and empirically accelerates convergence without degrading fidelity. 
For multimodal joint training, a generate-then-align three-stage schedule further improves stability and limits cross-task interference.
%
Experiments show that \Ours~achieves 2× faster convergence in training loss and up to 4× faster convergence in validation, while maintaining state-of-the-art performance on standard motion understanding and motion generation benchmarks.

{
}

{
}



\end{abstract}

%% file: sections/introduction.tex
\vspace{-8pt}
\section{Introduction}
\vspace{-8pt}
\label{intro}

Multimodal large language models (MLLMs) have recently achieved rapid progress in understanding and generation across text, images~\citep{team2024chameleon,wu2024janus,zhou2024transfusion}, 
audio~\citep{agostinelli2023musiclm,copet2023simple,liu2024audioldm}, and video~\citep{kondratyuk2023videopoet,zhang2023videollama,zhang2024video}.
Built on the strong semantic priors and in-context learning capabilities of pretrained LLMs, these models capture long-range dependencies and compositional structure,
enabling few-shot transfer and controllable citepacross modalities ~\cite{alayrac2022flamingo,chowdhery2023palm,dong2023dreamllm,li2023blip2,touvron2023llama}.
 Toward Unified Motion–Language Modeling.
 While most prior work has focused mainly on text-driven motion synthesis~\citep{shafir2023priormdm,tevet2022motionclip,mdm2022human,tevet2022human,chen2023mld,zhang2024motiondiffuse}, unified motion–language models for both understanding and generation remain comparatively underexplored.
 Pursuing both tasks in a single model demands representations and training strategies that respect the distinctive statistics of human motion without sacrificing the reasoning benefits of language models.

 Tokenizing motion into a fixed codebook, typically via VQ-based models, facilitates integration with Transformer-based LMs ~\citep{chuan2022tm2t,zhang2023t2mgpt,zhang2024motiongpt}, however inevitably introduces quantization error, attenuating high-frequency components and degrading semantic-physical consistency. 
 More importantly, treating motion as "language" ~\citep{jiang2023motiongpt,wang2023image,siyao2022bailando} 
 overlooks the gap between symbolic sequences and continuous trajectories 
 ~\citep{wang2025bridgingcontinuousdiscretetokens}.
 Consequently, cross-modal alignment often remains at a symbolic level and struggles to capture the fine-grained kinematics demanded by nuanced linguistic semantics. 
 In practice, limited codebook capacity and training coverage further constrain realism and controllability.

Recent MLLMs tend to process multiple modalities within a single backbone and attach modality-specific heads and supervision~\citep{park2025unifiedframeworkmotionreasoning,team2024chameleon,zhang2024large,zhou2024transfusion}.
However, jointly optimizing multimodal objectives induces gradient interference and loss-scale mismatch, 
which increases hyperparameter sensitivity, destabilizes training, and can erode language competence
~\citep{driess2023palm,kendall2018multi,tsimpoukelli2021multimodal}.
Moreover, 
forcing distinct modalities into a shared space
erodes modality-specific information and inductive biases, causing negative transfer.
For \textbf{robust, controllable motion–language modeling}, a method is needed that (i) adopts representations respect the continuous nature of human movement and (ii) explicitly balances multimodal, multi-objective training. 
Addressing these representational and optimization bottlenecks is, therefore, key to advancing unified motion understanding and generation.

\textbf{Continuous Motion Latent Space }
Our approach first replace the motion tokens with a continuous, low-dimensional latent representation learned by a pretrained motion VAE~\citep{chen2023mld}.
By ‘continuous’ we mean:
(i) the latents are real-valued vectors rather than discrete code indices, and (ii) the VAE induces a smooth latent manifold in which nearby points correspond to gradually varying motions.
Compared with VQ-based tokenization, this 
latent space is perceptually aligned with the original trajectories yet computationally compact, 
avoiding quantization artifacts and preserves high-frequency micro-dynamics for efficient and stable motion synthesis.

\textbf{Diffusion Bridge within the LLM Framework }
Autoregressive generation and cross-entropy–based supervision in LLMs presumes discrete token targets and is therefore ill-suited to continuous motion latents. 
Conditioned on the LLM’s hidden states, we further attach a lightweight diffusion head that perform denoising directly in the motion latent space to predict motion VAE latents, which the motion decoder then converts into motion sequences. 
Operating in a low-dimensional latent domain with a relatively small expert, this diffusion scheme bridges the gap between LLM hidden states and motion latents while only brings little overhead in both training and inference.

\textbf{Bimodal Architecture }
Following Mixture-of-Transformers (MoT)~\citep{liang2024mot}, we treat human motion as a second modality and introduce a motion branch symmetric to the language backbone. 
The two independent branches interact via shared attention layers, yet retain modality-specific embeddings and allow each module to be guided by its own objective.
This bimodal design mitigates interference between modalities and preserves each modality’s structure, thereby enabling high-quality motion understanding and generation within a unified framework. 

\textbf{Three-Stage Training }
To effectively model motion branch under guidance of a pre-trained language model, we design a three-stage training scheme.
First, we perform {Uni-task Pretraining}. with the text branch frozen, the motion branch is pre-trained on text-to-motion generation.
Next, in {Cross-Modal Alignment}, motion-to-text and motion prediction objectives are introduced to align two branches. 
Finally, all parameters are optimized in {Joint Fine-Tuning}.


We summarize our contributions as follows:
(i) Latent diffusion for motion.
Unlike quantization-based pipelines~\citep{zhang2023t2mgpt,zhang2024motiongpt},
we integrate latent diffusion~\citep{stable_diffusion,chen2023mld} into autoregressive backbone via
a diffusion head, bridging the continuous motion motion with the next-token prediction framework for higher-fidelity and diverse synthesis. 
(ii) Architecture and training.
We propose a bimodal motion-language framework with per-modal branches communicating through shared attention, reducing interference while preserving modality-specific intelligence.
A three-stage generate-then-align scheme further stabilizes joint training and curbs negative transfer.
(iii) Results and efficiency.
Under comparable settings, \Ours~achieves state-of-the-art performance on text-to-motion, motion-to-text, while reducing training time by approximately 2–3×.


%% file: sections/related.tex
\vspace{-8pt}
\section{Related Work}
\vspace{-6pt}
\label{relatedwork}
\textbf{Human Motion Modeling }
Early approaches leverage strong text encoders~\citep{li2022blip,radford2021clip,raffel2020exploringt5,sanh2019distilbert} to develop motion–language understanding/retrieval via shared embeddings~\citep{chuan2022tm2t,tevet2022motionclip,Yin_2024_CVPR} or contrastive learning ~\citep{chen2024motionllm,petrovich2023tmr}.
Recent methods
~\citep{athanasiou2024motionfix,cohan2024flexible,shafir2023priormdm,mdm2022human,chen2023mld,zhang2023remodiffuse,zhang2024large} advance text-to-motion generation with diffusion backbones ~\citep{ho2020denoising,song2020denoising}, either operating directly on raw motion sequence or reconstructing a VAE latent.
Working in a compressed latent space, LDMs~\citep{stable_diffusion} keeps training computationally cheaper and inference faster while maintaining synthesis quality.
In parallel, to fit next-token–prediction recipes in large language model (LLM), several works discretize motion with VQ-VAE ~\citep{esser2021taming,van2017vqvae} into token indices, enabling transformer-based generation~\citep{zhang2023t2mgpt,zhang2024motiongpt,zhang2025motion}.
However, quantization induces  approximation error and a "symbolic–continuous mismatch" that attenuates fine-grained kinematics and limits controllability, 
while refined tokenization such as residual VQ (RVQ)~\citep{guo2024momask} and post-training schemes~\citep{wang2025bridgingcontinuousdiscretetokens} only partially allieviate these issues and cannot fundamentally avoid the numerical and semantic discontinuities induced by tokenization.
Accordingly, we adopt an approach that interface language models directly with unquantized VAE latent representations.

Human motion modeling has evolved from task-specific designs to \textbf{unified frameworks for multimodal understanding and generation}. 
Recent motion models~\citep{jiang2023motiongpt, park2025unifiedframeworkmotionreasoning,wang2024motiongpt2,wu2025mgmotionllm} adopt single-stream LM backbones with discretized motion tokens to support bidirectional text-motion mapping, and have been attended to other modalities such as music~\citep{luo2024m3gpt,you2024momu}.
In parallel, language-centric multimodal frameworks extend LLMs beyond text via lightweight adapters and cross-attention conditioning, offering a generic recipe for vision- and audio-grounded reasoning~\citep{alayrac2022flamingo,li2023blip2,copet2023simple,liu2023audioldm}.
Works such as NExT-GPT~\citep{wu2024nextgpt} and Janus~\citep{wu2024janus,ma2024janusflow} employ pretrained encoders/decoders to map inputs into modality-specific latent spaces and attach adapters for flexible multimodal generation.
In vision-language, Chameleon~\citep{team2024chameleon} and Transfusion~\citep{zhou2024transfusion} discretize or encode images into token sequences to support interleaved text-image training, while Show-o~\citep{xie2024show} and Fuyu~\citep{fuyu-8b} employ masked or causal attention for joint reasoning. 
Despite their versatility, single-stream architectures often suffer from cross-modal interference, 
limiting scalability and robustness. 
Even with carefully tuned objectives,
newly introduced modalities can disrupt existing representations, 
underscoring the challenge of preserving modaliti-specific capability while scaling to new domains.

\textbf{Mixture-of-Experts and Multi-Stream Architecture } address these limitations by routing inputs to modality-specific experts while maintaining a shared fusion interface ~\citep{alayrac2022flamingo,li2023blip2,tsimpoukelli2021multimodal}.
This separation reduces gradient interference between modalities, and enables branch to be guided by its own objective~\citep{liu2021conflict,sener2018multi}, and simplifies the introduction of new modalities. 
These insights motivate hybrid strategies~\citep{cho2024discord,shi2025lmfusionadaptingpretrainedlanguage,wang2025bridging} that combine discrete and continuous representations and decouple modality-specific encoders with \emph{minimal} modification on the LLM backbone, thereby enhancing alignment and expressiveness.
Mixture-of-Transformers (MoT)~\citep{liang2024mot} 
instantiates this idea with modality-specific Transformer experts coupled through shared attention, facilitating modular training and reducing interference 
when incorporating new modalities. 
Guided by these observations, we adopt a MoT-style architecture that isolates motion representation learning while leveraging the language competence of pretrained LMs~\citep{bai2023qwen,radford2019gpt2}.

%% file: sections/method.tex
\vspace{-6pt}
\section{Method}
\label{sec:method}
\vspace{-8pt}

\begin{figure}[t]
    \centering
    \includegraphics[width=\linewidth]{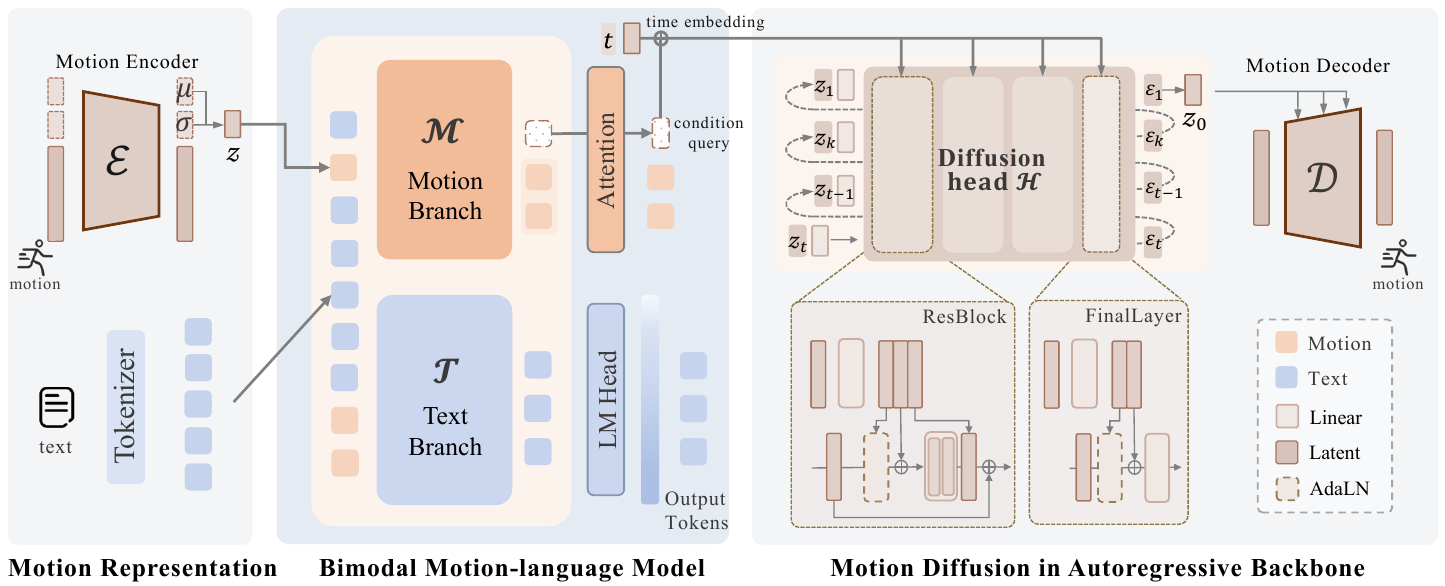}
  \vspace{-10pt}
    \caption{
    \Ours~introduces hybrid motion-language model that takes motion as a second modality and processes the data through a new branch, with cross-modal attention mechanism to communicate with text branch (\cref{sec:mot}). We leverage a  VAE network for continuous motion representation (\cref{sec:vae}), and design separate training objective for each modality (\cref{sec:diff}).
    }
    \label{fig:pipe}
  \vspace{-12pt}
\end{figure}

To couple motion understanding and generation into language-centric LLMs, 
we observe that although discretization in prior unified systems ~\citep{jiang2023motiongpt, wang2024motiongpt2, wu2025mgmotionllm} facilitates reuse of text-style training and inference pipelines, it inevitably removes fine-grained details and complicates optimization.
Moreover, single-stream backbones exacerbate cross-modal interference and yield imbalanced training. 
We circumvent these limitations by representing motion in a continuous, perceptually faithful VAE latent space (\cref{sec:vae}) and adopting a hybrid motion–text backbone that processes the two streams separately while permitting controlled interaction via shared self-attention (\cref{sec:mot}). 
On top of this backbone, we attach a diffusion head conditioned on LLM hidden states to bridge language and motion latents, enabling bidirectional understanding and generation (\cref{sec:diff}). 
Finally, together with a three-stage training schedule (\cref{sec:train}), our \Ours~avoids quantization bottlenecks and improves training and inference efficiency while maintaining generation quality.

\vspace{-6pt}
\subsection{Motion Representation in Continuous Tokens}
\vspace{-4pt}
\label{sec:vae}


To align motion with the autoregressive generation paradigm of large language models (LLMs)~\citep{bai2023qwen,radford2019gpt2,raffel2020exploringt5,touvron2023llama}, previous approaches typically 
\emph{descretizes} motion with vector-quantized autoencoders~\citep{esser2021taming,van2017vqvae}, converting a $N$-length sequences into latents $y\in \mathbb{R}^{n \times d}$ and replace each vector $y^i\in \mathbb{R}^{d}$ by its nearest code $e_k$ from a learned $K$-entry codebook. 
The corresponding indexes $k^{i...n}$ then serve as tokens for LLM training ~\citep{zhang2023t2mgpt,jiang2023motiongpt,wang2024motiongpt2}.
While they integrate cleanly with standard next-token prediction objectives in LLMs such as cross-entropy, quantization process inevitably introduces approximation error and disrupts motion continuity, weakening fine-grained dynamics and constraining controllability. 
~\citet{guo2024momask}, equiped with residual vector quantization (RVQ), leverages multiple codebooks whose decoded latents are summed to reduce information loss.
In parallel, ~\citet{wang2025bridgingcontinuousdiscretetokens} explores refined post-training tokenization. 
However, neither strategy fundamentally resolves the numerical and semantic discontinuities inherent to discretization.

In contrast, we adopt a {continuous} latent space learned by a motion VAE.
Given a $N$ frame motion sequence $m^{1:N}$, the encoder $\mathcal{E}$ map $m$ into a compact continuous latent vector $z \in \mathbb{R}^{d}$, and the decoder $\mathcal{D}$ reconstructs
$m^{1:M}=\mathcal{D}(z)=\mathcal{D}(\mathcal{E}(m^{1:M})$. 
The VAE is trained once with a reconstruction term (optionally including kinematic losses on pose and velocity) and a KL regularizer~\citep{kullback1951information} to prevent high-variance latents and promote a smooth manifold.
This compressed, continuous representation learns the inherent structure of $z$, and preserves subtle variations and maintains numerical and semantic continuity, while providing a compact domain in which our downstream generator operates.
Further details can be found in the supplement.

\vspace{-6pt}
\subsection{Bimodal Motion-Language Framework}
\label{sec:mot}
\vspace{-4pt}

To accommodate the distinct characteristics of language and motion while enabling efficient cross-modal interaction, we augment a decoder-only transformer backbone~\citet{radford2019gpt2} with a \emph{parallel} motion branch.
Unlike single-stream designs that merge all modalities into one pathway ~\cite{jiang2023motiongpt,wu2025mgmotionllm}, our architecture preserves modality-specific routes: a text branch $\mathcal{T}$ and a motion branch $\mathcal{M}$.
Each branch maintains its own embeddings, feed-forward blocks, and normalization, and information exchange occurs only in shared self-attention layers~\citep{alayrac2022flamingo,shi2025lmfusionadaptingpretrainedlanguage}.
The motion branch is initialized from scratch and trained primarily under its own objective, thereby capturing motion-specific inductive biases and reducing cross-modal interference during multimodal training~\citep{yu2020gradient,zhou2023intra}.



\textbf{Hybrid Sequence Route } 
As illustrated in \cref{fig:pipe}, given an input sequence $S=s^{1:k}$, each element is embedded either as a text embedding $\tau_i$ or as a motion latent $z_i$, with a routing indicator $\vartheta_i \in \{0,1\}$ dispatches them to $\mathcal{T}$ or $\mathcal{M}$.
The branches compute hidden states  $h_t$ and $h_m$ seperately, which are then reassembled in input order for shared self-attention layers.
This hybrid routing supports interleaved text–motion processing without collapsing modalities into a single embedding space, laying the foundation for high-quality, condition-aware generation.

\textbf{Interfaces for Continuous Motion Latents }
Because that the continuous motion representation~\citep{chen2023mld} does not rely on a tokenized vocabulary or codebook, the index-to-embedding lookup and softmax decoding employed for text cannot be reused for motion. 
We therefore introduce {motion-specific interfaces} that bridge continuous latents and transformer hidden states.
First, we augment the text vocabulary with a small set of motion-boundary/holder tokens (i.e. {\color{motionC} <som> <eom>}, {\color{motionC}<motion\_in>}, {\color{motionC}<motion\_out>}) to mark motion spans and I/O positions in interleaved sequences as in ~\citet{zhou2024transfusion}.
Second, a Motion Understanding Head (MUH) linearly maps motion latents into the Transformer's input embedding space for captioning and prediction.
Finally, a lightweight Motion Generation Head (MGH)
projects hidden states back to the VAE latent space via diffusion~\citep{ho2020denoising,stable_diffusion,chen2023mld}. 

\subsection{Motion Diffusion in Autoregressive Backbone}
\label{sec:diff}
\vspace{-4pt}
Continuous representations are inherently misaligned with the discrete nature of token-based generation in LLMs, and requires more sophisticated modeling to support generation.
Inspired by recent advances in diffusion-based generative modeling~\citep{mdm2022human,song2020denoising,stable_diffusion,mar},
we attach a lightweight diffusion module $\mathcal{H}$ to bridge this gap.
$\mathcal{H}$ predicts motion latents directly from the backbone’s hidden states, enabling integration of continuous motion representation within an autoregressive framework. 


\textbf{Diffusion Process }
Given a ground-truth motion sequence $x$, we obtain the target latent $z_0=\mathcal{E}(x)\in\mathbb{R}^d$ via the motion encoder $\mathcal{E}$.
We adopt a fixed forward noising process over $t\in\{1,\dots,T\}$ with Gaussian perturbations:
$z_t = \sqrt{\bar{\alpha}_t}\, z_0 + \sqrt{1-\bar{\alpha}_t}\,\epsilon$,
where $\epsilon\sim\mathcal{N}(0,I)$ and $\bar{\alpha}_t=\prod_{s=1}^{t}\alpha_s$ is the cumulative product of noise scheduling coefficients.
A time-aware denoiser $\mathcal{H}$ is conditioned on the Transformer hidden states from the motion stream (denoted $h_m$) and learns to reverse the diffusion process.
Conditioning is implemented via lightweight linear projections into the denoiser inputs, following recent practice~\citep{ho2022classifier,mar}.
We train $\mathcal{H}$ with the standard DDPM objective~\citep{ho2020denoising, song2020denoising}:
$\mathcal{L}_{\text{diff}}=
\mathbb{E}_{z_0,\,t,\,\epsilon}\!\left[\left\|\epsilon - \mathcal{H}(z_t, t, h_m)\right\|_2^2\right].
\vspace{-4pt}$

\textbf{Inference }
The text branch autoregressively generates tokens until a motion start marker {\color{motionC}{<som>}} is produced.
We then insert $K$ placeholder tokens (i.e., {\color{motionC}<motion\_out>})
to elicit the span-aligned hidden states $h_m^{i:i+K}$ in a single forward pass.
The diffusion head runs the reverse process conditioned on $h^{i:i+K}$ to sample the noise-free motion latent $\hat{z}_0$, which the VAE decoder $\mathcal{D}$ finally decoded to the raw motion sequence.
As in ~\citet{team2024chameleon}, generation then resumes in the text stream with a concatenated {\color{motionC}{<eom>}} until end-of-sequence.
Operating on compact motion latents, this diffusion head adds only minimal overhead during training and inference.

    

\subsection{Training Procedure}
\label{sec:train}
\vspace{-4pt}

\begin{figure}[t] 
    \centering
  \vspace{-5pt}
    \includegraphics[width=1.\linewidth]{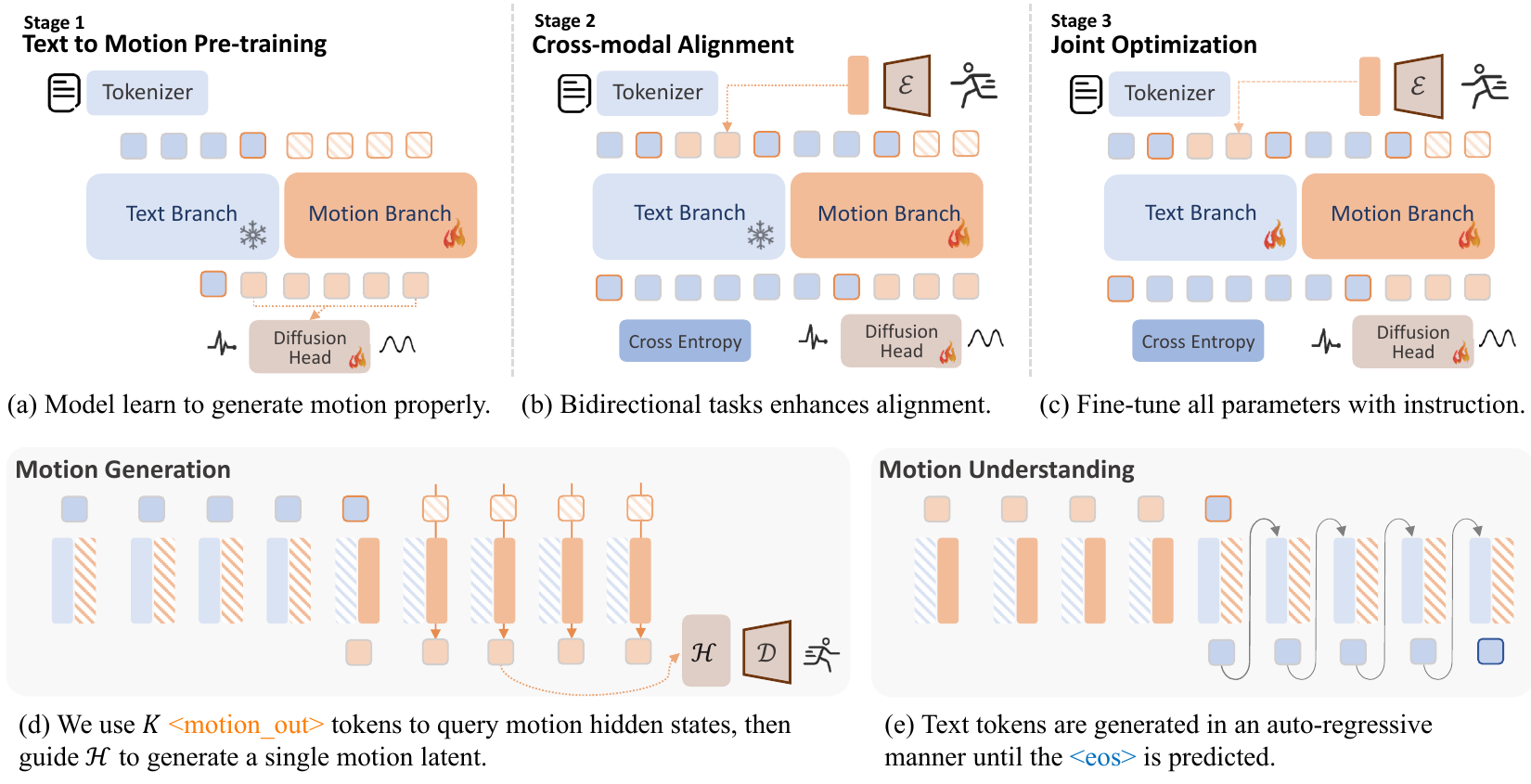}

  \vspace{-5pt}
    \caption{
    We propose a \textbf{three-stage alignment} strategy for our hybrid motion-language model:
    (a) The text branch is frozen, and only motion output is supervised.
    (b) Motion reasoning is introduced to further align the motion branch with language, with supervision on both modalities.
    (c) All modules are jointly fine-tuned with text branch unfrozen. 
    (d)(e) shows inference time behavior of two branches which process data only with the same modality tags (differtiated by colors). Each rectangle block represents for a whole text/motion branch, while shadowed ones denote inactive modules.
    Modalities are color-coded: {\color{textC} blue for text} and {\color{motionC} orange for motion}. 
    Shadowed orange squares represent {\color{motionC} \texttt <motion\_out>}, and orange-outlined squares indicate boundary tokens {\color{motionC} \texttt <som>} or {\color{motionC} \texttt <eom>}.
    }
  \vspace{-10pt}
    \label{fig:training}
\end{figure}

    
We first train a motion VAE to obtain a compact, continuous latent space, following prior works~\citep{stable_diffusion,chen2023mld}. The bimodal backbone then adopts a pretrained decoder-only LLM~\citep{radford2019gpt2} as the text branch $\mathcal{T}$. As described in \cref{fig:training}, the motion branch $\mathcal{M}$ is initialized from scratch and brought into alignment with $\mathcal{T}$ through a three-stage schedule.

\textbf{Stage I: Text-to-motion pretraining }
We begin by pretraining $\mathcal{M}$ on text-to-motion, while freezing$\mathcal{T}$.
This provides stable linguistic conditioning and biases the model toward motion-specific semantics.
In this stage, $\mathcal{M}$ conditions on the frozen language representations and is trained via diffusion, to synthesize VAE motion latentss, where diverse text–motion pairs encourages a rich and flexible mapping from language to the latent space~\citep{stable_diffusion,chen2023mld}.


\textbf{Stage II: Cross-Modal Alignment }
Keeping $\mathcal{T}$ frozen, we introduce additional objectives to couple understanding and generation. 
Concretely, training includes multiple tasks of text-to-motion (T2M), motion-to-text (M2T), and motion prediction.
Following instruction-style formulations in~\citet{jiang2023motiongpt}, these tasks are further presented as prompts covering generation, captioning, prediction, and inbetweening.
Multi-task optimization fosters bidirectional alignment without forcing a single shared representation and encourages motion representations that are semantically coherent with language features~\citep{alayrac2022flamingo,li2023blip2}.


\textbf{Stage III: Joint Fine-Tuning }
Finally, we unfreeze $\mathcal{T}$ and fine-tune all parameters via instruction tuning on a mixture of paired text–motion data and, optionally, text-only prompts~\citep{dai2023instructblip,liu2023visual,wei2021finetuned}.
Including text-only prompts can further improve language competence for downstream applications.

%% file: sections/experiment.tex
\vspace{-2pt}
\section{Experiments}
\vspace{-8pt}
\label{experiments}

We empirically validate \Ours, a dual-stream architecture for efficient, language-grounded multimodal motion understanding and generation, across motion-centric tasks.
Dataset configurations, evaluation metrics, and implementation details are summarized in \cref{sec:exp-detail}. 
Begin with analyzing optimization dynamics and inference efficiency via training loss and validation curves
(\cref{sec:exp-speed}), we then present controlled ablations that isolate the contributions of the continuous VAE motion representation and the bimodal design (\cref{sec:exp-abl}). 
Next, We benchmark \Ours~on text-to-motion generation and motion-to-text understanding, comparing against both specialized single-task methods and unified state-of-the-art systems (\cref{sec:exp-comp}). 
Finally, we ablate the proposed three-stage training scheme (\cref{sec:exp-stage}). 
Additional qualitative results are provided in the supplement.


\vspace{-4pt}
\subsection{Experimental Setup}
\vspace{-4pt}
\label{sec:exp-detail}

\textbf{Datasets } 
We train and evaluate our model on ~\citet{guo2022generating}, a large-scale benchmark for text-motion generation and understanding.
For comparison with prior works~\citep{chen2023mld,jiang2023motiongpt}, we adopt the 263-dim pose proposed in~\citet{guo2022generating}, which combines joint velocities/ positions/ rotations, and foot-contact signals,
following the standard data split.

\textbf{Evaluation Metrics }
We evaluate two tasks.
For the \emph{text-to-motion}, we follow the previous works~\citep{chuan2022tm2t,jiang2023motiongpt,chen2023mld,zhang2023t2mgpt} to report motion quality (FID), diversity (DIV and MM), and text-motion alignment (R-Precision and MMDist).
For \emph{motion-to-text}, we use both alignment metrics (R-Precision and MMDist) and linguistic metrics from NLP (Bleu~\citep{papineni2002bleu}, Rouge-L~\citep{lin2004rouge}, CIDEr~\citep{vedantam2015cider}, and BertScore~\citep{zhang2019bertscore}).
See \cref{sec:appendix:metrics} for metric definitions and computation details.




\textbf{Implementation Details}
Our framework comprises three main components: a motion VAE, a lightweight diffusion head, and a dual-stream backbone.
We adopt the Transformer-based motion VAE of~\citet{chen2023mld}, where both encoder and decoder consists of 9 layers and 4 heads with skip connections, 
producing a $1\times1\times256$ latent per motion sequence.
Our Diffusion Head $\mathcal{H}$ is implemented as a 3-layer MLP with ResBlock-style layers and hidden dimension 1024, following~\citet{mar}.
We train diffusion with a scaled linear noise schedule for 1000 denoising steps, while inference uses 100 steps by default.
The text and motion branches share the GPT-2 base configuration but use disjoint parameters, both are decoder-only with 12 Transformer layers, model dimension 768, and MLP dimension 3072, unless stated otherwise.
The text branch is initialized from a pretrained 124M GPT-2 checkpoint,
while the motion branch from scratch, yielding total 238M parameters.


\textbf{Training Protocol }
We use AdamW for all components, with a learning rate of $1 \times 10^{-4}$ for the motion backbone and $2 \times 10^{-4}$ for the diffusion head.
Training uses a mini-batch size of 32 on 2 NVIDIA RTX 3090 GPUs, with identical training/inference settings on HumanML3D~\citep{guo2022generating}.
The motion VAE is trained with a learning rate of $1\times10^{-4}$, batch size 256, over 150K iterations. 
The motion-language backbone is trained for 100k iterations in text-to-motion pretraining, followed by 300k iterations for cross-modal alignment.

\begin{figure}[t]
    \centering
\vspace{-4pt}
    \includegraphics[width=\linewidth]{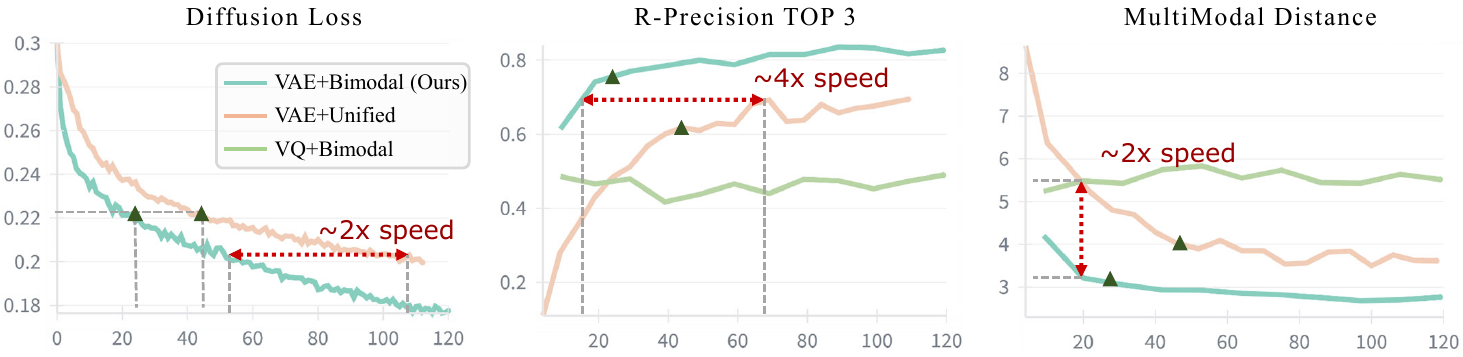}
    \caption{
    Training loss and validation curves on motion generation on HumanML3D for architecture variants of dual-stream and single-stream and representation variants of VAE and VQ latents.
    The right figures illustrate validation metrics of R-Precision TOP 3 (R@3$\uparrow$) and Multimodal Distance (MMDist$\downarrow$).
    Triangle markers indicate matched-loss checkpoints ($\sim$0.22).
    Our hybrid architecture with continuous motion representation helps accelerating convergence for about 2$\times$, as well as achieves better quality especially in early training stage.
    }

    \vspace{-5pt}
    \label{fig:exp:speedcurve}
\end{figure}

\vspace{-4pt}
\subsection{Training Efficiency with Bimodal Architecture and Continuous Latents}
\label{sec:exp-speed}
\vspace{-4pt}

To assess our two main design choices: (i) the dual-stream (bimodal) backbone and (ii) continuous representation from motion VAE, 
we analyze training dynamics on the text-to-motion task under identical settings on HumanML3D.
\cref{fig:exp:speedcurve} plots training loss and validation metrics over time for three variants of single-stream+VAE, dual-stream+VAE, and dual-stream+VQ.

We observe that
i) \textbf{Discrete VQ latents plateau at 
a lower quality ceiling.}
The VQ baseline (VQ + bimodal; green curve) reaches $\approx$0.5 R-Precision Top 3 (R@3) early in training and then saturates, yielding substantially lower R@3 and higher MultiModal Distance (MMDist) than the VAE-based counterparts. 
This likely stems from 
quantization-induced information loss~\citep{guo2024momask, wang2025bridging} and tokenization that disrupts the semantic continuity of motion.
ii) \textbf{Bimodal design accelerates optimization. }
Compared with single-stream backbone, the dual-stream variant reduces diffusion loss roughly 2× faster and sustains superior validation performance over the entire training trajectory, in terms of R@3 and MMDist.
iii) \textbf{Superior quality at matched optimization states. }
At comparable diffusion loss, the dual-stream model still leads. 
For instance, we mark by triangles a loss of $\approx$0.22,
reached at $\sim$20 epochs for dual-stream model and $\sim$40 epochs for the single-stream model, the former achieves higher R@3 and lower MMDist.

We attribute these effects to \textbf{modality-aware optimization}: the motion branch learns motion-specific semantics while the language branch focuses on textual cues. 
Decoupling the streams and supervising them separately mitigates gradient interference and avoids the representational compromises common in single-stream models~\citep{liu2021conflict,sener2018multi,yu2020gradient}.
Although a shared space might be expected to bring modalities "closer", in practice single-stream coupling can entangle modality structure and yield counterintuitive outcomes.
In summary, continuous VAE latents within a dual-stream backbone strike a favorable efficiency–fidelity trade-off, enabling high-quality motion synthesis with reduced training time.

\vspace{-4pt}
\subsection{Comparisons}
\label{sec:exp-comp}
\vspace{-4pt}
By modeling human motion as a second modality alongside language, our bimodal motion-language model supports both text-to-motion (T2M) and motion-to-text (M2T).
We report results for two settings: (i) single-task models trained specifically for target task (\Ours$^\dagger$) 
, and (ii) a unified model (\Ours) trained on both tasks with the three-stage scheme 
.



\begin{table}[t]
 \centering
 \scriptsize
 \caption{    
   Evaluation of text-guided motion generation on HumanML3D ~\citep{Guo_2022_CVPR_humanml3d}.
 Rows are grouped by training tasks: \emph{Gen. only} for generation-only and \emph{Gen. \& Und.} for both.
 \textit{Real} is obtained by ground-truch motions, 
 and $\rightarrow$ indicate values closer to \textit{Real} are desirable. 
 $\dagger$ marks our single-task model trained for 200 epochs, and 
 \Ours~is a three-stage model trained with unified tasks.
 }
\vspace{-2pt}
\resizebox{\linewidth}{!}{%
\begin{threeparttable}
 \begin{tabular}{@{}llccccccc@{}}
 \toprule
    Types & Methods & R@1 & R@2 & R@3 & FID$\downarrow$ & MMDist$\downarrow$
    & Diversity$\rightarrow$ & MModality$\uparrow$ \\
  \midrule
    &
    Real & $0.511$ & $0.703$ & $0.797$ & $0.002$ & $2.974$ & $9.503$ & - \\
  \midrule
    \multirow{4}{*}{Gen. only}&
    T2M-GPT~\cite{zhang2023t2mgpt} & $0.491$ & $0.68$ & $0.775$ 
    & $0.116$ 
    & $3.118$ & $9.761$ & $1.856$ 
    \\
    &
    DiverseMotion~\cite{lou2023diversemotion} & $0.515$ & $0.706$ & $0.802$ 
    & \underline{0.072} 
    & $2.941$ & $9.683$ & $1.869$ 
    \\
    &
    MoMask~\cite{guo2024momask} & $0.521$ & $0.713$ & $0.807$ 
    & \textbf{0.045}
    & $2.958$ & \underline{9.620} & $1.241$
    \\
    &
    \textbf{\Ours}$\dagger$ 
    & \underline{0.533} & 0.731 & \underline{0.826} & 0.239& \underline{2.797} &9.688 & 1.560 \\
    
  \midrule
    \multirow{4}{*}{Gen. \& Und.}&
    TM2T~\cite{chuan2022tm2t} & $0.424$ & $0.618$ & $0.729$ 
    & $1.501$
    & $3.467$ & $8.589$ & \textbf{2.424}
    \\
    &
    MotionGPT~\cite{jiang2023motiongpt} & $0.492$ & $0.681$ & $0.733$ 
    & $0.232$ 
    & $3.096$ & \textbf{9.528} & $2.00$ 
    \\
    &
    MoTe~\cite{wu2024mote} & $0.548$ & \underline{0.737} & $0.825$ 
    & $0.075$ 
    & $2.867$ & - & \underline{2.399}
    \\
    &
    \textbf{\Ours}
    & \textbf{0.553} & \textbf{0.747} & \textbf{0.837} & 0.208 & \textbf{2.725} & 9.700 & 1.018 \\    
    
 \bottomrule
 \end{tabular}
 
\end{threeparttable}
} 
 \label{tab:comp-t2m}
\end{table}

\textbf{Text-to-Motion Generation } 
\label{sec:exp-t2m}
The text-to-motion task involves generating realistic and diverse motion sequences conditioned on natural language descriptions.
We train a single-task generator \Ours$\dagger$ for 200 epochs on HumanML3D and compare against recent methods~\citep{guo2022generating,chuan2022tm2t,lou2023diversemotion,jiang2023motiongpt,chen2023mld,zhang2023t2mgpt,zhang2023remodiffuse,guo2024momask,wu2024mote,li2024lamp,zhang2025motion}.
Following~\citet{guo2022generating}, each evaluation is repeated 20 times and reported with 95\% confidence intervals.
As shown in \cref{tab:comp-t2m}, \Ours$\dagger$ matches or exceeds generation-only baselines~\citep{zhang2023t2mgpt,lou2023diversemotion,guo2024momask} 
on alignment metrics (R@k, MMDist), with competitive FID and diversity.
The unified \Ours~is performs on par with or better than recent unified systems~\citep{chuan2022tm2t,jiang2023motiongpt,wu2024mote}.
\cref{sec:appendix:quantitative} provides additional comparisons.


\textbf{Motion-to-Text Understanding }
\label{sec:exp-m2t}
The motion-to-text task involves understanding motion sequences and generating semantically appropriate textual descriptions.
We train a single-task captioner \Ours$^\dagger$ for 100 epochs and compare with recent SOTA~\citep{chuan2022tm2t, jiang2023motiongpt, li2024lamp, wu2024mote}.
Following~\citet{jiang2023motiongpt}, we evaluate on the raw ground truth texts using the TM2T protocol~\citep{chuan2022tm2t}. 
Results in \cref{tab:comp-m2t} show that both \Ours$^\dagger$ and \Ours~
achieve strong retrieval performance and language metrics.
Notably, we observe a marked reduction in Multimodal Distance (MMDist), 
indicating effective motion–language alignment under the dual-stream design.

\begin{table}[t]
  \centering
\vspace{-2pt}
  \caption{
 Comparison of motion captioning on HumanML3D ~\citep{Guo_2022_CVPR_humanml3d}, evaluation follows \citep{chuan2022tm2t}. 
 \Ours~$\dagger$ denotes our single-task captioning model trained for 100 epochs, and 
 \Ours~is an unified model trained on both tasks with the three-stage scheme (\cref{sec:train}).
 Both variants achieve R@k on par with recent state of the art,
 and surpass the GT metrics.
 }
\vspace{-2pt}
\resizebox{\columnwidth}{!}{%
\begin{threeparttable}
  \begin{tabular}{@{}lccccccccc@{}}
  \toprule
     Methods & R@1 & R@2 & R@3 & MMDist $\downarrow$ 
     & Bleu@1 $\uparrow$ & Bleu@4 $\uparrow$ 
     & Rouge $\uparrow$ & Cider $\uparrow$ & BertScore $\uparrow$ \\
  \midrule
    Real     & 0.523 & 0.725 & 0.828 & 2.901 & -    & -    & -    & -    & -    \\
  \midrule
    
    TM2T~\citep{chuan2022tm2t}    
    & 0.516 & -     &  0.823 & 2.935 & {48.9} & 7.00 & {38.1} & 16.8 & 32.2 \\
    
    MotionGPT~\citep{jiang2023motiongpt}    
    & 0.543 & -     & 0.827 & 2.821 & 48.2 & {12.5} & 37.4 & {29.2} & {32.4} \\
    
    LaMPM2T~\citep{li2024lamp}
    & 0.547 & - & 0.831 & 2.808 & 47.8 & {13.04} & 37.1 & 28.9 & {32.7}\\
    MoTe~\citep{wu2024mote} 
    & \textbf{0.577} & - & \textbf{0.871} & {2.649} & 46.7 & 11.15 & 37.4 & \textbf{31.5} & 30.3 \\
    \textbf{\Ours$\dagger$} 
    & 0.553 & 0.756 & 0.853 & \underline{2.524} & \underline{56.363} & \underline{17.661} & \underline{44.997} & \underline{30.980} & \textbf{35.850}
    \\
    \textbf{\Ours} 
    & \underline{0.573} & \textbf{0.773} & \underline{0.864} & \textbf{2.426} & \textbf{59.083} & \underline{19.412} & \textbf{46.173} & 28.721 & \underline{35.231}
    \\
    
 \bottomrule
 \end{tabular}
\end{threeparttable}
}
  \label{tab:comp-m2t}
\vspace{-10pt}
\end{table}

   

\vspace{-4pt}
\subsection{Ablation Studies}
\label{sec:exp-abl}
\vspace{-4pt}

This section reports quantitative ablations.
In contrast to training-curve analysis in \cref{sec:exp-speed}, we evaluate final test-set performance on both text-to-motion (T2M) and motion-to-text (M2T).
First, we assess the contributions of a dual-stream backbone and continuous VAE motion latents by varying one factor at a time.
Then, we analyze the proposed three-stage training schedule and quantify its effects
We also examine the impact of hidden-state processing in the Diffusion Head $\mathcal{H}$ and the use of classifier-free guidance (CFG). See \cref{sec:appendix:abl} for more detailed experiments.

\textbf{Model Design }
\label{sec:exp-abl-arch}
\cref{tab:abl-arch} summarized test-set results on HumanML3D~\citep{Guo_2022_CVPR_humanml3d}.
We evaluate T2M and M2T separately and compare four variants obtained by crossing architecture (single- vs. dual-stream) with representation (discrete VQ vs. continuous VAE). Under the same evaluation protocol, replacing VAE with VQ or replacing a dual-stream backbone (Bimodal) with a single-stream one (Unified) consistently degrades performance on both tasks.
Notably, changing the architecture change to Bimodal yields larger gains on M2T, whereas changing the representation to VAE yields larger gains on T2M.
This task-dependent sensitivity is consistent with \cref{sec:exp-speed}: decoupling streams mitigates cross-modal interference and benefits semantic-level alignment, while continuous latents reduce quantization loss and improve synthesis fidelity for motion generation.



{
}

\begin{table}[t]
 \centering
 \scriptsize
 \caption{
    Component ablations on HumanML3D for representation choice and architecture design.
   \emph{Unified} denotes a single-stream backbone, where one branch is shared by text and motion, as employed in ~\citet{jiang2023motiongpt}, and \emph{Bimodal} denotes a dual-stream backbone described in \cref{sec:mot}. 
   VQ and VAE indicate discrete and continuous motion latents, respectively.
   For each configuration we train separate models for motion generation (T2M) and motion captioning (M2T) under the same protocol and report test-set metrics.
   All variants share the same GPT-2-style branch and hyperparameters.
   and training is run for 100 epochs on M2T and 200 epochs on T2M.
   Best and second-best results are highlighted in \textbf{bold} and \underline{underline}.
 }
\vspace{-2pt}
\resizebox{\columnwidth}{!}{%
\begin{threeparttable}

 \begin{tabular}{@{}lcccccccc@{}}
 \toprule
    \multirow{2}{*}{Settings} & \multicolumn{4}{c}{Text-to-Motion} & \multicolumn{4}{c}{Motion-to-Text}
    \\ \cmidrule(lr){2-5} \cmidrule(lr){6-9}
    & R@1 $\uparrow$ & R@3 $\uparrow$ & MMDist $\downarrow$ & FID $\downarrow$
    & R@1 $\uparrow$ & R@3 $\uparrow$ & MMDist $\downarrow$ & BertScore $\uparrow$$\uparrow$
    \\
  \midrule
    Real & $0.511^{\pm0.003}$ & $0.797^{\pm0.002}$ & $2.974^{\pm0.008}$  & $0.002^{\pm0}$
    & 0.523 & 0.828 & 2.901 & -  
    \\
  \midrule
    Unified+VQ &
    $0.237^{\pm0.003}$ & $0.435^{\pm0.003}$ & $5.684^{\pm0.018}$ & $\underline{0.403}^{\pm0.014}$ 
    & - & - & - & -
    \\
    Unified+VAE &
    $\underline{0.501}^{\pm0.003}$ & $\underline{0.792}^{\pm0.002}$ & $\underline{2.841}^{\pm0.011}$ & $0.489^{\pm0.017}$ 
    & 0.234 & 0.426 & 5.976 & 16.197
    \\
    Bimodal+VQ & $0.300^{\pm0.005}$ & $0.532^{\pm0.02}$ & $4.937^{\pm0.077}$ & $0.454^{\pm0.078}$ 
    & $\underline{0.379}$ & $\underline{0.702}$  & $\underline{3.545}$ & $\underline{18.085}$
    \\    
    Bimodal+VAE 
    & $\textbf{ 0.533}^{\pm0.002}$ & $ \textbf{0.826}^{\pm0.003}$ 
    & $\textbf{2.797}^{\pm0.007}$ & $ \textbf{0.239}^{\pm0.008}$ 
    & $\textbf{0.553}$  & $\textbf{0.853}$  & $\textbf{2.524}$ & $\textbf{35.850}$
    \\
 \bottomrule
 \end{tabular}
 \begin{tablenotes}
 {
     \footnotesize
     \item T2M with \textit{Bimodal+VQ} and \textit{Unified+VQ} is extended to 400 epochs to approach convergence. \\
     M2T results for \textit{Unified+VQ} is not reported are omitted because performance remained unevaluable after more than 400 training epochs.
     }
 \end{tablenotes}
\end{threeparttable}


 }
 \vspace{-10pt}
 \label{tab:abl-arch}
\end{table}

\textbf{Training Stage }
\label{sec:exp-stage}
We ablate the three-stage schedule in \cref{sec:train}, including 100k iters on text-to-motion pretraining (SI), 300k iters on cross-modal alignment (SII), and 50k iters on joint fine-tuning (SIII), and evaluate on both T2M and M2T.
Results are summarized in \cref{tab:abl-training}.
SI already yields strong generation and provides a motion-specialized initialization.
Optimization in SII confers M2T capability and, importantly, further improves T2M, by $-0.10$ on FID and $-0.2$ on MMD.
indicating that explicit alignment benefits both directions.
Without extra text-only supervision, SIII adds small additional gains on M2T while preserving T2M, serving as a light joint refinement rather than a substitute for S2.
As shown in the last row of \cref{tab:abl-training}, a two-stage model that \emph{omits SI} keeps M2T largely intact but markedly degrades T2M, underscoring the role of S1 in learning motion-specific features.
Overall, the full three-stage schedule provides the best trade-off, 
delivering reliable generation and captioning with well-aligned motion–language representations.
Additional variants are reported in \cref{sec:appendix:stage}, including experiments with an unfrozen text branch.


\begin{table}[t]
\centering
\caption{Ablation on training-scheme.
Enabled stages are marked with \Checkmark, and colors encode the text branch {\color{textC}updated} or \textbf{frozen}.
Best results are \textbf{bold} and second best are \underline{underlined}.
}
 \vspace{-2pt}
\resizebox{.86\columnwidth}{!}{%
    \begin{tabular}{@{}ccc cccccc@{}}
  \toprule
    \multirow{2}{*}{Stage I} & 
    \multirow{2}{*}{Stage II} & 
    \multirow{2}{*}{Stage III} & 
    \multicolumn{3}{c}{Text-to-Motion}& 
    \multicolumn{3}{c}{Motion-to-Text} 
    \\ 
    \cmidrule(lr){4-6} \cmidrule(lr){7-9} 
    &&& R TOP3 $\uparrow$ & FID $\downarrow$ & MMDist $\downarrow$ & R TOP1$\uparrow$ & Bleu@4$\uparrow$ & Bert$\uparrow$ 
    \\ 
  \midrule
    \Checkmark & \XSolidBrush & \XSolidBrush
    & 0.826 & 0.239 & 2.797
    & - & - & -
    \\
    \Checkmark &  \Checkmark  & \XSolidBrush
    & 0.831 & 0.215 & 2.755
    & 0.571 & 18.328 & 33.993
    \\
    
    \Checkmark &  \Checkmark  & {\color{textC} \Checkmark}
    & 0.837 & 0.208 & 2.725
    & 0.573 & 19.412 & 35.231
    \\
    \XSolidBrush & \Checkmark & {\color{textC}\Checkmark}
    & 0.772 & 0.325 & 3.108
    & 0.573 & 18.277 & 35.546
    \\
  \bottomrule
    \end{tabular}%
}
\label{tab:abl-training}
\vspace{-10pt}
\end{table}

%% file: sections/discussion.tex
\vspace{-6pt}
\section{Discussion}
\vspace{-8pt}

To address quantization-induced degradation and cross-modal interference in multi-objective training,
we present \textbf{\Ours}, a dual-stream motion-language framework that unifies motion understanding and generation while preserving modality-specific inductive biases.
By encoding motion as continuous VAE latents and generating in latent space with a lightweight diffusion head, the model avoids quantization artifacts and improves synthesis fidelity.
The dual-stream Transformer with shared attention enables controlled bidirectional exchange, which strengthens text–motion alignment, reduces cross-modal interference, and empirically accelerates single-task convergence without degrading quality.
For joint training of understanding and generation, a generate-then-align three-stage training schedule further stabilizes optimization and mitigates cross-task interference.

\textbf{Limitations and Failure Cases }
Fine-grained control cam fail on directional cues (e.g., left/right).
Because the current VAE yields a single latent per sequence, segment-level composition and local semantic alignment for long motions are not explicitly supported.
Generalization to out-of-domain descriptions is constrained by data coverage.
Potential remedies include incorporating diverse text-only corpora in the final alignment stage, adopting stronger language backbones, and exploring hierarchical or segment-wise latent representations to enable compositional control.
\textbf{Future Work } We will scale training to (i) larger, more diverse datasets, (ii) develop controllable motion with local semantic alignment and segment-level for long-horizon generation, and (iii) evaluate the framework with stronger language models and larger-scale training regimes to assess efficiency and robustness.

{
}

{


}

{

}

%% file: sections/appendix.tex
\renewcommand\thesection{\Alph{section}}
\renewcommand*{\theHsection}{appedix.\thesection}
\setcounter{section}{0}
\setcounter{figure}{4}
\setcounter{table}{3}
\setcounter{equation}{1}

This appendix provides qualitative comparison results (\cref{sec:appendix:qualitative}), disccusion of continuous/ discrete motion representation (\cref{sec:appendix:representation}), additional quantitative results and ablation (\cref{sec:appendix:abl}) on motion branch size and connection type, motion supervision scheme, training stages. 
We also 
provide more implementation details in \cref{sec:appendix:detail}.
Please note our examination of 
metrics report on TMR evaluator (\cref{sec:appendix:tmr}), 
analysis on bimodal architecture  (\cref{sec:appendix:mot-arch}),
and ablation on our training scheme (\cref{sec:appendix:stage}).


\textbf{Website \& Video }
A supplementary website provides visualizations of quantitative results, motion data, and demonstration videos. 
A standalone video is also available on the website and at \path{supp/website_video/static/videos/MotionGPT3/_Video.mp4}, showcasing (i) text-to-motion comparisons, (ii) motion-captioning comparisons, and (iii) additional results on motion generation and captioning.

\textbf{Code }
Example code files are available in the supplementary materials, which cover the training and evaluation processes of our \Ours~models, along with several example results.

\vspace{-4pt}
\section{Discrete Token vs Continuous Token}
\label{sec:appendix:representation}
\vspace{-4pt}

\textbf{Reconstruction Task } 
We compare continuous latents from MLD VAE~\citep{chen2023mld} with discrete latents from VQ-VAE~\citep{jiang2023motiongpt} for motion reconstruction under identical settings.
As shown in \cref{tab:appendix:vae_vq_recon}, VQ-VAE yields higher errors on MPJPE, PAMPJPE, ACCL, and APE/AVE for root, trajectory, pose, joints, indicating reduced fidelity and temporal smoothness relative to the continuous VAE.
This gap is expected: quantization maps a continuous motion manifold to a finite codebook and turns motion modeling into token classification, introducing unavoidable approximation error and information loss.
In practice, many distinct frames collapse to the same code (one-to-many mapping), yielding ambiguous reconstructions and frame-wise noise that harms smoothness.


\begin{table}[hbp]
\label{tab:appendix:vae_vq_recon}
 \centering
    \caption{Reconstruction performance of a continuous VAE~\citep{chen2023mld}
    versus a discrete VQ-VAE~\citep{jiang2023motiongpt}.
    The VQ-VAE shows consistently higher errors, consistent with information loss introduced by quantized encoding and decoding.
    \cref{sec:appendix:metrics} presents the metric definitions.
    }
\vspace{-3pt}
 \resizebox{\columnwidth}{!}{%
 \begin{tabular}{l ccc cccc cccc}
 \toprule
     \multirow{2}{*}{Method} & \multirow{2}{*}{MPJPE} & \multirow{2}{*}{PAMPJPE} & \multirow{2}{*}{ACCL} 
     & \multicolumn{4}{c}{APE} & \multicolumn{4}{c}{AVE} \\
     \cmidrule(lr){5-8}  \cmidrule(lr){9-12} 
     & & & & root & traj & pose & joints & root & traj & pose & joints \\

 \midrule
    VAE & 43.906 & 31.356 & 5.93   
    & 0.0581 & 0.0504 & 0.0277 & 0.0619 
    & 0.0179 & 0.0177 & 0.0012 & 0.0185 
    \\
    VQ  & 46.828 & 33.668 & 7.629  
    & 0.0829 & 0.0804 & 0.0316 & 0.0930 
    & 0.0240 & 0.0239 & 0.0015 & 0.0253 
    \\
 \bottomrule
    \end{tabular}
}
\end{table}

\textbf{On Discrete VQ Latents } 
Recent work ~\citet{guo2024momask} addresses the limitations of codebook capacity using  hierarchical Residual Vector Quantization (RVQ) with separate predictors for base and residual tokens. ~\citet{wang2025bridgingcontinuousdiscretetokens} combines quantized and continuous latents via post-training quantization.
~\citet{cho2025discorddiscretetokenscontinuous} 
introduces a diffusion-based decoder that progressively maps discrete tokens back to continuous raw motions, improving fidelity and smoothness, and uses the symmetric Jerk Percentage Error (sJPE) to detect under-reconstruction and frame noise.
Despite these advances, discrete pipelines remain prone to expressiveness bottlenecks and token-induced jitter. We further evaluate VQ and VAE latents within dual-stream architecture under single-task training. The results (\cref{tab:appendix:vq_vae}) consistently favor continuous representations for both generation and understanding.
While discrete codes facilitate token-based modeling, continuous representations better capture fine-grained dynamics and motion continuity, achieving higher alignment and synthesis quality with fewer epochs.

\begin{table}[hb]
  \centering   
    \caption{
    Discrete (VQ) vs. continuous (VAE) motion representations under single-task training.
    We report both T2M on R@1, FID, MMDist, DIV and M2T on R@3, BLEU@1/4, ROUGE.
    With fewer training epochs, VAE-variants achieve stronger alignment and better quality.
    VQ requires extended training of 399 epochs while still remains behind on most alignment and language scores.
    }
    \vspace{-3pt}
 \scriptsize
\resizebox{\columnwidth}{!}{%
\begin{threeparttable}
  \begin{tabular}{@{}l ccccc ccccc@{}}
  \toprule
    \multirow{2}{*}{Representation} & 
    \multicolumn{5}{c}{Text-to-Motion}& 
    \multicolumn{5}{c}{Motion-to-Text}
    \\ \cmidrule(lr){2-6} \cmidrule(lr){7-11}
    & epoch
    & R@1$\uparrow$ 
    & FID $\downarrow$ 
    & MM Dist $\downarrow$ 
    & DIV$\rightarrow$ 
    & epoch
    & R@3$\uparrow$ 
    & Bleu@1$\uparrow$ 
    & Bleu@4$\uparrow$ 
    & Rouge$\uparrow$
    \\ 
  \midrule
    VQ   & 199
    & $0.258$  
    & $0.542$ & $5.364$ & $9.274$
    & 99 
    & 0.765 & 47.043 & 7.234 & 39.244
    \\
     & 399 
    & $0.300$ 
    & $0.454$ & $4.937$ & $9.626$
    & 199 
    & 0.752 & 41.579 & 6.304 & 35.746

    \\
  \midrule
    VAE & 199 
    & $0.525$ 
    & $0.191$ & $2.667$ & $10.095$
    & 99 
    & 0.859 
    & 50.707 & 8.383 & 38.225 
    \\
  \bottomrule
    \end{tabular}
\end{threeparttable}
}
    \vspace{-4pt}
    \label{tab:appendix:vq_vae}
\end{table}

\begin{figure}[htbp]
 \vspace{-6pt}
 \centering
    \includegraphics[width=.96\linewidth]{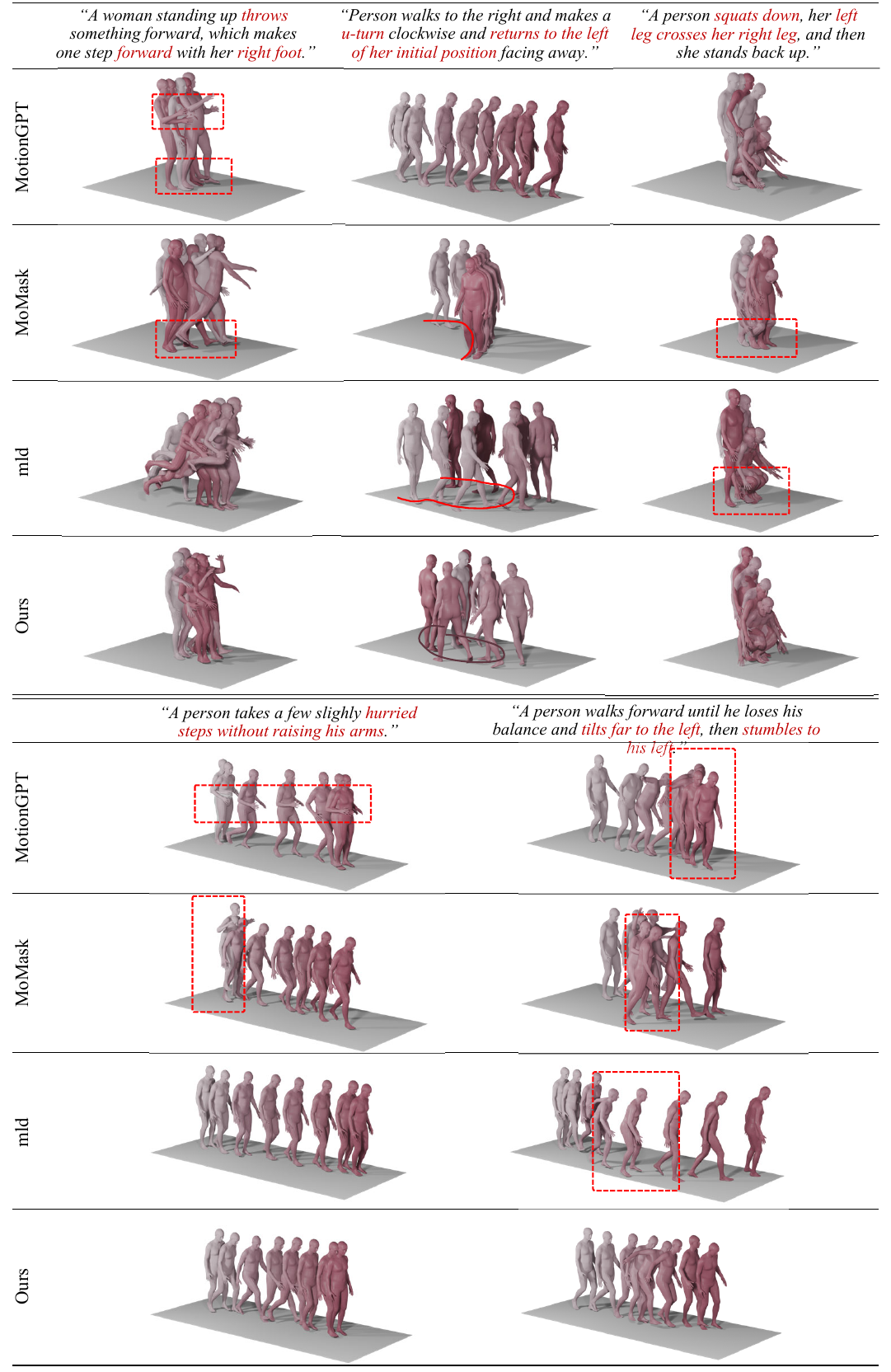}
 \caption{Qualitative comparison of text-driven motion generation on HumanML3D~\citep{Guo_2022_CVPR_humanml3d}.
 Baselines (~\citep{jiang2023motiongpt,guo2024momask,chen2023mld}) are run with their official released checkpoints.
 {\color{red}Red annotations} (text, boxes, curves) highlight prompt–motion mismatches.
  Our bimodal motion-language framework 
  yields motions that with closer correspondence to the textual prompt and smoother temporal coherence.
 }
  \vspace{-10pt}
    \label{fig:appendix:t2m}
\end{figure}

 \begin{figure}[ht]
     \centering
     \includegraphics[width=.95\linewidth]{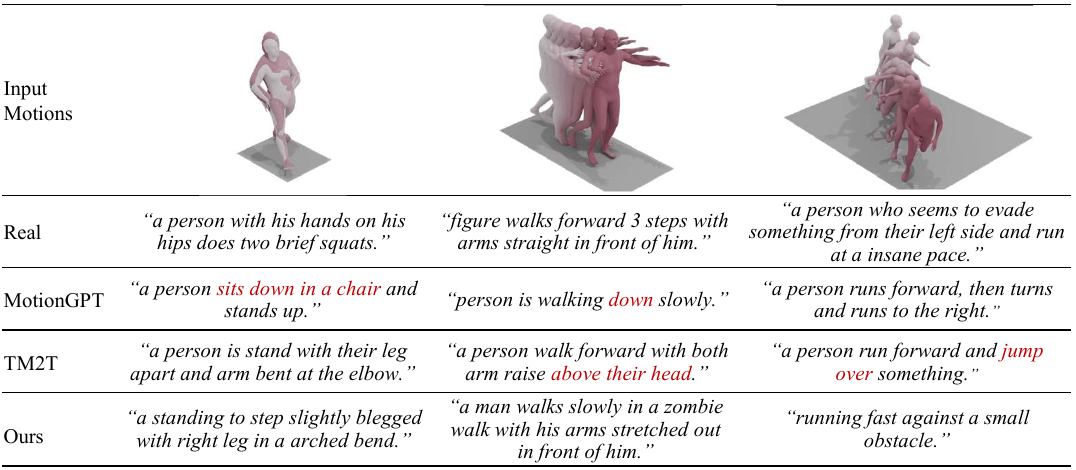}
  \vspace{-3pt}
     \caption{Example results on motion caption. The misalignment is hilighted with {\color{red} red}.
     }
  \vspace{-3pt}
     \label{fig:appendix:m2t}
 \end{figure}
 
\begin{figure}[!ht]
    \centering
    \includegraphics[width=.95\linewidth]{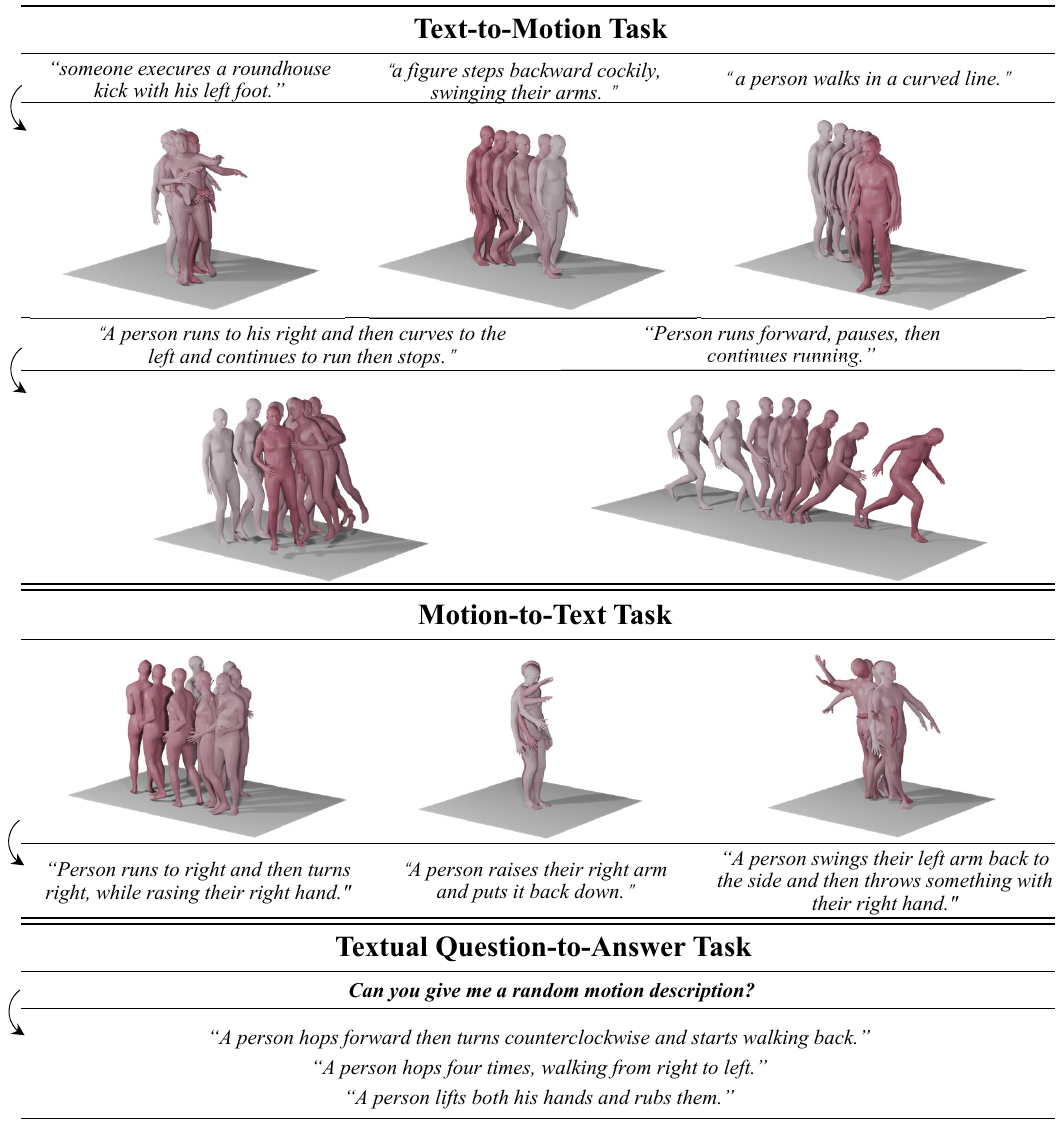}
  \vspace{-3pt}
    \caption{Gallery for the results of \Ours. 
    Top: text-to-motion generation. Middle: motion-to-text captioning. Bottom: textual question answering about motion.
    Examples are produced by our unified model trained with instruction-based objectives (three-stage scheme). 
    Animated visualizations are provided in the supplementary video.
    } 
  \vspace{-5pt}
    
    \label{fig:appendix:more_rs}
\end{figure}

\textbf{Generation Task (T2M) } 
With the same 200 training epochs, the VAE representation delivers substantially better quality than VQ, with 0.525 vs. 0.258 on R@1 and 2.667 vs. 5.364 on MMDist, while maintaining competitive diversity(DIV).
Extending VQ to 399 epochs reduces FID to 0.454,
however, it still lags in alignment, with R@1 0.300 and MMDist 4.937,
indicating a lower quality ceiling for discrete codes.

\textbf{Understanding Task (M2T) } 
At matched training over 99 epochs, VAE attains stronger retrieval and language scores than VQ, with 0.859 vs. 0.765 on R@3 and 50.707 vs. 47.043 on BLEU@1, while ROUGE is comparable.
Prolonged VQ training to 199 epochs unexceptedly reduces performance across all language scores with R@3 dropping by 0.13 and BLEU@1 by 5.464,
suggesting optimization instability and poorer generalization under the discrete setting.


\section{Qualitative Results}
\label{sec:appendix:qualitative}
\vspace{-4pt}

We provide qualitative comparisons for text-to-motion (T2M) \cref{fig:appendix:t2m}, motion-to-text (M2T) \cref{fig:appendix:m2t}, and a multi-task gallery \cref{fig:appendix:more_rs}. 
To ensure fair visualization, we use a fixed list of prompts from the HumanML3D test split; all clips are rendered at 20 FPS with identical camera and skeleton settings. 
Baselines are run with the their official checkpoints. 
Overall, \Ours~exhibits stronger smoother transitions and text-motion alignment.
Discrete-code variants tend to show token-induced frame noise and temporal drift, whereas single-stream models can produce semantically inconsistent motions on complex prompts. 
Additional examples and animations are provided in the supplementary video.


\begin{table}[!h]
 \centering
 \vspace{-1pt}
 \caption{
 Comprehensive comparison of text-to-motion generation on HumanML3D~\citep{Guo_2022_CVPR_humanml3d}. 
 We report generation-only models (Gen. only) here, and visualize unified dual-task models (Gen. \& Und.) in ~\cref{fig:comp-t2m-unified}.
 Real denotes ground-truth statistics; arrows ($\rightarrow$) indicate that values closer to Real are desirable.
  $\dagger$ marks our single-task model trained for 200 epochs, and \Ours~is the unified three-stage model.
  Best and second-best results are \textbf{bold} and \underline{underlined}.
 }
 \scriptsize
\resizebox{\columnwidth}{!}{%
 \begin{threeparttable}
 \begin{tabular}{@{}lccccccc@{}}
 \toprule
    Methods & R@1 & R@2 & R@3 & FID$\downarrow$ & MMDist$\downarrow$ & DIV$\rightarrow$ & MModality$\uparrow$ \\
 \midrule
    Real & $0.511^{\pm0.003}$ & $0.703^{\pm0.003}$ & $0.797^{\pm0.002}$ & $0.002^{\pm0}$ & $2.974^{\pm0.008}$ & $9.503^{\pm0.065}$ &  \\
 \midrule
    T2M~\citep{guo2022generating} 
    & $0.457^{\pm0.002}$ & $0.639^{\pm0.003}$ & $0.74^{\pm0.003}$ 
    & $1.067^{\pm0.002}$ 
    & $3.34^{\pm0.008}$ & $9.188^{\pm0.002}$ & $2.09^{\pm0.083}$ 
    \\
    
    MLD~\citep{chen2023mld}
    & $0.481^{\pm0.003}$ & $0.673^{\pm0.003}$ & $0.772^{\pm0.002}$ 
    & $0.473^{\pm0.013}$ 
    & $3.169^{\pm0.01}$ & $9.724^{\pm0.082}$ & $\underline{2.413}^{\pm0.079}$ 
    \\
    
    MotionDiffuse~\citep{zhang2024motiondiffuse} 
    & $0.491^{\pm0.001}$ & $0.681^{\pm0.001}$ & $0.782^{\pm0.001}$ 
    & $0.63^{\pm0.001}$ 
    & $3.113^{\pm0.001}$ & $\underline{9.410}^{\pm0.049}$ & $1.553^{\pm0.042}$ 
    \\
    
    T2M-GPT~\citep{zhang2023t2mgpt} 
    & $0.491^{\pm0.003}$ & $0.68^{\pm0.003}$ & $0.775^{\pm0.002}$ 
    & $0.116^{\pm0.004}$ 
    & $3.118^{\pm0.011}$ & $9.761^{\pm0.081}$ & $1.856^{\pm0.011}$ 
    \\
    
    ReMoDiffuse~\citep{zhang2023remodiffuse}
    & $0.51^{\pm0.005}$ & $0.698^{\pm0.006}$ & $0.795^{\pm0.004}$ 
    & $0.103^{\pm0.004}$ 
    & $2.974^{\pm0.016}$ & $9.018^{\pm0.075}$ & $1.795^{\pm0.043}$ 
    \\
    
    DiverseMotion~\citep{lou2023diversemotion} 
    & $0.515^{\pm0.003}$ & $0.706^{\pm0.002}$ & $0.802^{\pm0.002}$ 
    & $0.072^{\pm0.004}$ 
    & $2.941^{\pm0.007}$ & $9.683^{\pm0.102}$ & $1.869^{\pm0.089}$ 
    \\
    
    MoMask~\citep{guo2024momask} 
    & $0.521^{\pm0.002}$ & $0.713^{\pm0.002}$ & $0.807^{\pm0.002}$ 
    & $\underline{0.045}^{\pm0.002}$ 
    & $2.958^{\pm0.008}$ & ${9.620}^{\pm0.064}$ & $1.241^{\pm0.04}$ 
    \\
    
    MotionAnything~\citep{zhang2025motion} 
    & $\underline{0.546}^{\pm0.003}$ & $\underline{0.735}^{\pm0.002}$ & $\underline{0.829}^{\pm0.002}$ 
    & $\textbf{0.028}^{\pm0.005}$ 
    & ${2.859}^{\pm0.01}$ & $\textbf{9.521}^{\pm0.083}$ & $\textbf{2.705}^{\pm0.06}$
    \\
    \textbf{\Ours$\dagger$}
    & ${0.533}^{\pm0.002}$ & ${0.731}^{\pm0.002}$ & $ {0.826}^{\pm0.003}$  & $0.239^{\pm0.008}$ 
     & $ \underline{2.797}^{\pm0.007}$ & $9.688^{\pm0.107}$ & $1.560^{\pm0.052}$ \\
    
    \textbf{\Ours}
    & $\textbf{0.553}^{\pm0.003}$ & $\textbf{0.747}^{\pm0.002}$ & $\textbf{0.837}^{\pm0.003}$ & $0.208^{\pm0.006}$ & $\textbf{2.725}^{\pm0.008}$ & $9.700^{\pm0.096}$ & $1.018^{\pm0.038}$ \\
    
 \bottomrule
 \end{tabular}
 \begin{tablenotes}
    \footnotesize
    \item[$\dagger$] We train our model on single T2M task for 200 epochs.
 \end{tablenotes}
 \end{threeparttable}
}
 \label{tab:comp-t2m-all}
\end{table}

\newpage

\section{Additional Experiments}
\label{sec:appendix:abl}
\vspace{-4pt}

This section provides supplementary evaluations that complement the main results. 
First, we report a comprehensive comparison including additional generation-only (Gen. only) and unified dual-task baselines (Gen. \& Und.) (\cref{sec:appendix:quantitative}) and further assess text-to-motion with the TMR retrieval evaluator (\cref{sec:appendix:tmr}). 
We then analyze design choices of the Cross-Modal Attention (CMA) (\cref{sec:appendix:mot-arch}), examine scaling effects of both language and motion branches (\cref{sec:appendix:abl-size}),
and analyze the diffusion-based supervision used for motion generation (\cref{sec:appendix:vae-abl}). 
Fianlly, we evaluate the effectiveness of the three-stage training strategy (\cref{sec:appendix:stage}).

\subsection{Quantitative Results}
\label{sec:appendix:quantitative}
\vspace{-4pt}


\cref{tab:comp-t2m-all} reports the full text-to-motion results on HumanML3D, grouped by training regime (generation-only vs. unified dual-task).
Notably, evaluated on the HumanML3D dataset by T2M evaluator~\citep{guo2022generating}, recent models consistently achieve very high scores, and several recent approaches
~\citep{guo2024momask,zhang2025motion,li2024lamp,wu2024mote,wu2025mgmotionllm}, including ~\Ours, achieve scores
even above those of the ground-truth data (\emph{Real}).

\newpage
\begin{figure}[h]
    \centering
    \vspace{-6pt}
    \includegraphics[width=0.8\linewidth]{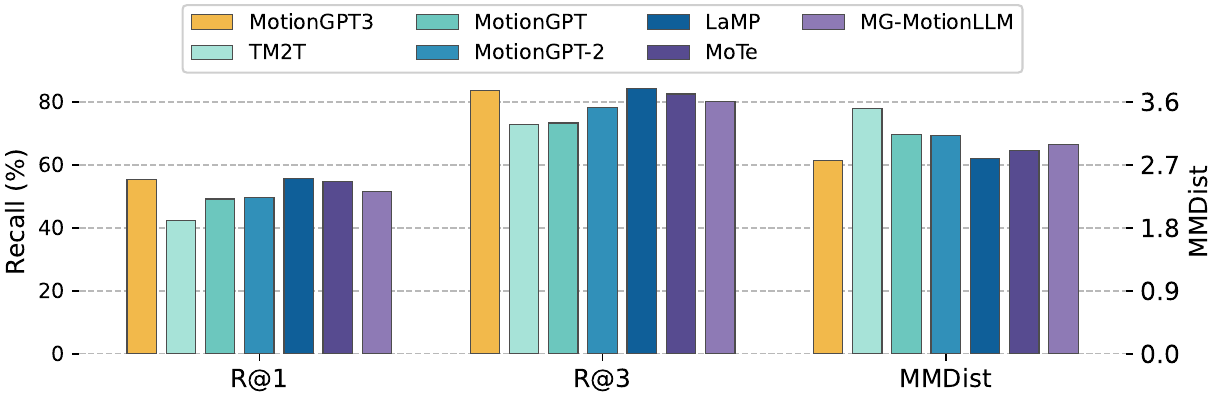}
    \vspace{-8pt}
    \caption{Comparison on text-to-motion, with recent approaches trained with unified tasks(Gen. \& Und.). Our model performs betther than recent unified models: MotionGPT~\citep{jiang2023motiongpt}, TM2T~\citep{chuan2022tm2t}, MotionGPT-2~\citep{wang2024motiongpt2}, MG-MotionLLM~\citep{wu2025mgmotionllm}, MoTe~\citep{wu2024mote}, and comparable with LaMP~\citep{li2024lamp}.
    }
    \vspace{-8pt}
    \label{fig:comp-t2m-unified}
\end{figure}

Generally speaking, high evaluation scores indicates that the generated motions can correspond well with the text, and approximate high-fidelity motions.
\textbf{However}, considering that the GT embeddings effectively represent an upper bound for matching scores, the practical significance of differences among methods achieving near or above GT performance might be limited.
This reflects a limitation of the T2M evaluator, where the metrics are computed in a learned embedding space which relies on contrastive learning on HumanML3D. Methods that overfit to that space can saturate the proxy and even surpass the ground-truth reference, without a proportionate improvement in motion fidelity.
Hence, differences among methods near or above the Real line are hard to interpret.


\vspace{-2pt}
\subsection{Evaluation with TMR}
\label{sec:appendix:tmr}
\vspace{-4pt}

For a more nuanced assessment, we further evaluate with the TMR retrieval framework~\citep{petrovich2023tmr}, under four gallery protocols:
\textbf{All}, entire test set,
\textbf{All with threshold}, gallery items whose textual similarity to the query exceeds a fixed threshold,
\textbf{Dissimilar}, a 100-pair subset with mutually distant texts, and
\textbf{Small batches}, mini-batch of size 32 to mimic~\citet{guo2022generating} setting.
We report text–motion retrieval with R@1/2/3/5/10 and MedR, results are summarized in \cref{tab:appendix-tmr}.

Not all methods in \cref{tab:comp-t2m-all} release TMR results or checkpoints, so the comparison is limited to publicly available models.
Within this scope, although inferior to ~\citet{li2024lamp} on T2M metrics (\cref{fig:comp-t2m-unified}), our method achieves significantly stronger retrieval under the TMR evaluator and improves substantially over the T2M baseline~\citet{guo2022generating} across protocols.
These findings suggest that our model achieves more robust cross-modal alignment rather than overfitting to a specific evaluator.

\begin{table}[h]
    \centering
\vspace{-4pt}
    \caption{
    Retrieval on HumanML3D with the TMR evaluator~\citep{petrovich2023tmr}.
    We report R@1/2/3/5/10 and MedR for text-motion retrieval under the four official protocols: (a) All, (b) All with threshold, (c) Dissimilar subset, and (d) Small batches (see \cref{sec:appendix:tmr} for definitions).
    Results for TEMOS~\citep{petrovich22temos}, T2M~\citep{guo2022generating}, and TMR~\citep{petrovich2023tmr} are taken from the TMR paper.
    LaMP~\citep{li2024lamp} is reported only for (d).
    MotionGPT and \Ours~are evaluated with the released checkpoints using the official TMR code.
    \Ours~attains strong performance across protocols.
    }
    \vspace{-4pt}
\resizebox{\columnwidth}{!}{%
    \begin{tabular}{@{}ll|cccccc|cccccc@{}}
  \toprule
    \multirow{2}{*}{Protocol} & \multirow{2}{*}{Methods} 
    & \multicolumn{6}{c|}{\text Text-motion retrieval}& \multicolumn{6}{c}{\text Motion-text retrieval} \\
    & & R@1$\uparrow$ & R@2$\uparrow$ & R@3$\uparrow$ & R@5$\uparrow$ & R@10$\uparrow$ & MedR$\downarrow$ 
    & R@1$\uparrow$ & R@2$\uparrow$ & R@3$\uparrow$ & R@5$\uparrow$ & R@10$\uparrow$ & MedR$\downarrow$ 
    \\
    \cmidrule{1-2}
    \cmidrule(lr){3-8}
    \cmidrule(lr){9-14}
  All 
      & TEMOS & 2.12 & 4.09 & 5.87 & 8.26 & 13.52 & 173.00 
      & 3.86 & 4.54 & 6.94 & 9.38 & 14.00 & 183.25 \\
      & T2M & 1.80 & 3.42 & 4.79 & 7.12 & 12.47 & 81.00 
      & 2.92 & 3.74 & 6.00 & 8.36 & 12.95 & 81.50 \\
      & TMR & 5.68 & 10.59 & 14.04 & 20.34 & 30.94 & 28.00 
      & 9.95 & 12.44 & 17.95 & 23.56 & 32.69 & 28.50 \\
      & MotionGPT & 7.16 & 12.50 & 15.85 & 21.53 & 30.20 & 38.00 
      & 11.31 & 13.91 & 19.39 & 24.13 & 31.80 & 36.25 \\
      \rowcolor{gray!20}
      \cellcolor{white} & \Ours 
      & \textbf{9.60} & \textbf{17.36} & \textbf{22.45} & \textbf{30.43} & \textbf{41.06} & \textbf{17.00} 
      & \textbf{14.90} & \textbf{18.20} & \textbf{24.43} & \textbf{31.32} & \textbf{40.72} & \textbf{17.50} \\
  \midrule
  All with threshold 
      & TEMOS & 5.21 & 8.22 & 11.14 & 15.09 & 22.12 & 79.00 
      & 5.48 & 6.19 & 9.00 & 12.01 & 17.10 & 129.00 \\
      & T2M & 5.30 & 7.83 & 10.75 & 14.59 & 22.51 & 54.00 
      & 4.95 & 5.68 & 8.93 & 11.64 & 16.94 & 69.50 \\
      & TMR & 11.60 & 15.39 & 20.50 & 27.72 & 38.52 & 19.00 
      & 13.20 & 15.73 & 22.03 & 27.65 & 37.63 & 21.50 \\
      & MotionGPT & 14.32 & 21.01 & 25.94 & 33.39 & 43.84 & 15.00 
      & 14.42 & 16.83 & 22.70 & 27.69 & 35.06 & 30.50 \\
      \rowcolor{gray!20}
      \cellcolor{white} & \Ours 
      & \textbf{20.73} & \textbf{27.03} & \textbf{34.03} & \textbf{42.66} & \textbf{52.97} & \textbf{9.00} 
      & \textbf{19.34} & \textbf{22.40} & \textbf{29.40} & \textbf{36.91} & \textbf{46.30} & \textbf{13.00} \\
  \midrule
  Dissimilar subset 
      & TEMOS & 33.00 & 42.00 & 49.00 & 57.00 & 66.00 
      & 4.00 & 35.00 & 44.00 & 50.00 & 56.00 & 70.00 & 3.50 \\
      & T2M & 34.00 & 48.00 & 57.00 & 72.00 & 84.00 
      & 3.00 & 34.00 & 47.00 & 59.00 & 72.00 & 83.00 & 3.00 \\
      & TMR & 47.00 & 61.00 & 71.00 & 80.00 & 86.00 & 2.00 
      & 48.00 & 63.00 & 69.00 & 80.00 & 84.00 & 2.00 \\
      & MotionGPT & 51.00 & 64.00 & 71.00 & 74.00 & 80.00 & 1.00 
      & 53.00 & 62.00 & 68.00 & 76.00 & 81.00 & 1.00 \\
      \rowcolor{gray!20}
      \cellcolor{white} & \Ours 
        & \textbf{68.00} & \textbf{77.00} & \textbf{85.00} & \textbf{92.00} & \textbf{95.00} & \textbf{1.00} 
        & \textbf{63.00} & \textbf{73.00} & \textbf{83.00} & \textbf{89.00} & \textbf{93.00} & \textbf{1.00} \\
  \midrule
  Small batches 
      & TEMOS & 40.49 & 53.52 & 61.14 & 70.96 & 84.15 & 2.33 
      & 39.96 & 53.49 & 61.79 & 72.40 & 85.89 & 2.33 \\
      & T2M & 52.48 & 71.05 & 80.65 & 89.66 & 96.58 & 1.39 
      & 52.00 & 71.21 & 81.11 & 89.87 & 96.78 & 1.38 \\
      & TMR & 67.16 & 81.32 & 86.81 & 91.43 & 95.36 & 1.04 
      & 67.97 & 81.20 & 86.35 & 91.70 & 95.27 & 1.03 \\
      \rowcolor{yellow!40}
      \cellcolor{white} & LaMP 
      & 67.18 & 81.90 & 87.04 & {92.00} & {95.73} & - 
      & 68.02 & 82.10 & 87.50 & 92.20 & 96.90 & - \\
      & MotionGPT & 58.07 & 69.91 & 74.34 & 79.17 & 86.36 & 1.18 
      & 58.71 & 69.64 & 74.36 & 79.45 & 86.02 & 1.16 \\
      \rowcolor{gray!20}
      \cellcolor{white} & \Ours 
      & \textbf{74.25} & \textbf{86.70} & \textbf{91.29} & \textbf{94.82} & \textbf{97.35} & \textbf{1.00} 
      & \textbf{74.00} & \textbf{86.86} & \textbf{91.04} & \textbf{94.62} & \textbf{97.35} & \textbf{1.00} \\
  \bottomrule
    \end{tabular}
}
    \label{tab:appendix-tmr}
  \vspace{-6pt}
\end{table}

\newpage
\subsection{Motion Branch with Cross-Modal Connection} 
\label{sec:appendix:mot-arch}

Our hybrid model allows asymmetric capacities for the text and motion branches and supports different patterns of cross-modal information exchange. 
In this section we focus on \emph{where} to place cross-modal attention (CMA) in the backbone for the text-to-motion task, keeping all other factors fixed (\cref{tab:appendix:mot-attn}).
Ablation on branch capacity is deferred to \cref{sec:appendix:abl-size}.

We explore several CMA schedules that differ only in the layers where cross-modal connections are enabled(\cref{tab:appendix:mot-attn}).
Across paired settings with the same spacing pattern, shifting the CMA blocks to later layers typically improves generation quality and distribution similarity to the ground-truth (i.e., lower FID and MMDist, higher R-Precision).
This is most evident in $B_{1}$ v.s. $B_{2}$: both use uniformly spaced CMA with identical count, $B_{1}$ enable CMA from the first to the second-last layer, while $B_{2}$ shifts them by one layer to span the second through the last layer. 
Despite the minor offset, $B_{2}$ achieves noticeably better scores.
The same tendency appears in $B{3}$–$B{4}$, $C{1}$–$C{3}$, and $D_{1}$-$D_{2}$, where the distribution pattern is matched but the CMA positions differ.

\begin{figure}[h]
    \centering
    \includegraphics[width=.76\linewidth]{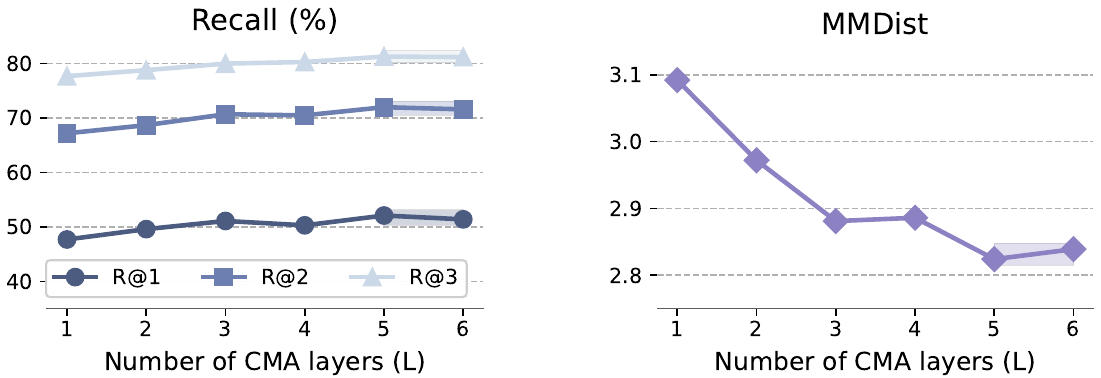}
    \vspace{-2pt}
    \caption{Ablation on the number of cross-modal attention (CMA) layers for T2M on HumanML3D. CMA is enabled in the last $L$ layers ($L\in{1,\dots,6}$).
Performance improves as $L$ increases up to 5 layers, then shows slight degradation at 6, indicating a \textbf{non-monotonically pattern}.}
    \label{fig:appendix-attn-last}
\end{figure}

We further enable CMA in the last $L$ layers and sweep $L$ (\cref{fig:appendix-attn-last}).
Increasing $L$ from 2 to 5 generally improves quality, reflected by lower FID scores and MMDist and higher R-Precision.
However, the trend is \textbf{non-monotonic}: we observe a slight drop at $L=6$  in R-Precision, relative to $L=5$, may suggesting that late but not ubiquitous CMA is preferable.

\newcommand{\ca}[1]{\color{brown}{#1}}
\newcommand{\cb}[1]{\color{cyan}{#1}}
\newcommand{\cc}[1]{\color{Violet}{#1}}
\newcommand{\cn}[1]{{#1}}
\newcommand{\arrowcn}{\ensuremath{{\cn \leftrightharpoons}}}
\newcommand{\arrowca}{\ensuremath{{\ca \leftrightharpoons}}}
\newcommand{\arrowcb}{\ensuremath{{\cb \leftrightharpoons}}}
\newcommand{\arrowcc}{\ensuremath{{\cc \leftrightharpoons}}}

\begin{table}[hb]
  \hfill
 \caption{Cross-modal attention(CMA) configurations used in ablation.
 (a) Layer-wise CMA schedules for configurations A–D across the 12-layer backbones.
 Within each branch, the symbol \arrowcn~ marks a cross-modal attention (CMA) operation at that layer, blanks empty indicate intra-modal attention only.
 (b) Schematics diagrams for configurations $A$, $B_1$, $C_1$.
 }
 \begin{subfigure}{\textwidth}
  \begin{minipage}[b]{0.55\linewidth}
    \centering
  \vspace{2pt}
\resizebox{\linewidth}{!}{
    \begin{tabular}{@{} l cccccccccccc @{}}
    \toprule
    Model & 0 & 1 & 2 & 3 & 4 & 5 & 6 & 7 & 8 & 9 & 10 & 11 \\
    \midrule
    $A$  & \arrowcn & \arrowcn & \arrowcn & \arrowcn & \arrowcn & \arrowcn & \arrowcn & \arrowcn & \arrowcn & \arrowcn & \arrowcn & \arrowcn \\
    $B_{\cn{1}}$  & \arrowcn &  & \arrowcn &  & \arrowcn &  & \arrowcn &  & \arrowcn &  & \arrowcn &  \\
    $B_{\cn{2}}$  &  & \arrowcn &  & \arrowcn &  & \arrowcn &  & \arrowcn &  & \arrowcn &  & \arrowcn \\
    $B_{\cn{3}}$  & \arrowcn & \arrowcn & \arrowcn & \arrowcn & \arrowcn & \arrowcn &  &  &  &  &  &  \\
    $B_{\cn{4}}$  &  &  &  &  &  &  & \arrowcn & \arrowcn & \arrowcn & \arrowcn & \arrowcn & \arrowcn \\
    $C_{\cn{1}}$ & \arrowcn & \arrowcn & \arrowcn & \arrowcn &  &  &  &  &  &  &  &  \\
    $C_{\cn{2}}$ &  &  &  &  & \arrowcn & \arrowcn & \arrowcn & \arrowcn &  &  &  &  \\
    $C_{\cn{3}}$ &  &  &  &  &  &  &  &  & \arrowcn & \arrowcn & \arrowcn & \arrowcn \\
    $D_{\cn{1}}$ & \arrowcn &  &  &  &  &  &  &  &  &  &  &  \\
    $D_{\cn{2}}$ &  &  &  &  &  &  &  &  &  &  &  & \arrowcn \\
    \bottomrule
    \end{tabular}
}
    \tabcaption{Layer-wise CMA settings of each experiment in \cref{tab:appendix:mot-attn}.}
  \vspace{-8pt}
  \end{minipage}
  \hfill
  \begin{minipage}[b]{0.42\linewidth}
      \centering
    \includegraphics[width = .94\linewidth]{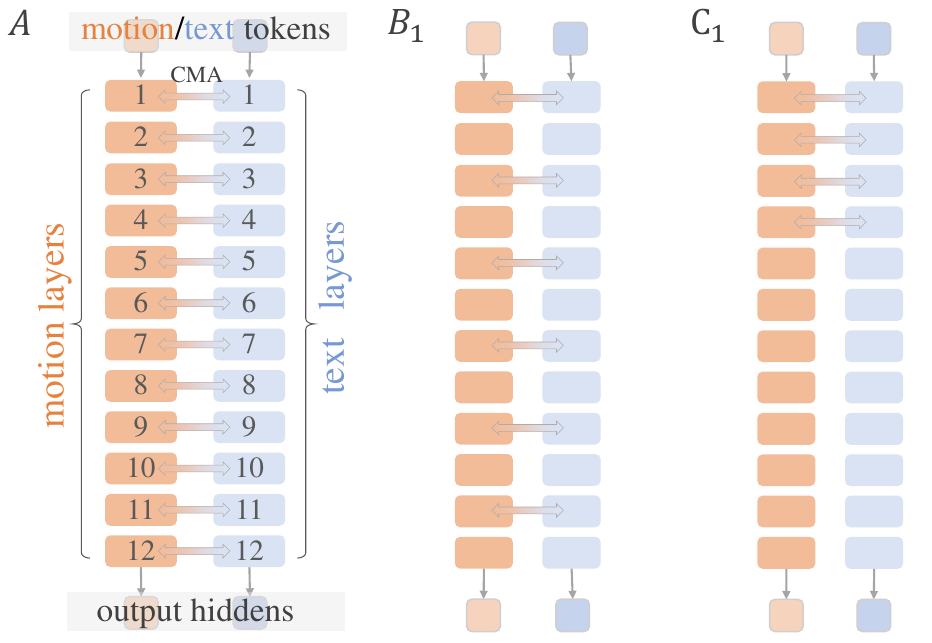}
  \vspace{-6pt}
    \figcaption{Connection visualization of $A$, $B_1$ and $C_1$.}
  \vspace{-8pt}
  \end{minipage}
 \end{subfigure}
\label{tab:appendix-attn-lab}
\end{table}

\begin{table}[h]
\centering
\caption{
Quantitative results for several CMA settings on T2M with 200K training iterations, settings visualized in \cref{tab:appendix-attn-lab}.
The text branch is pretrained GPT-2 (124M) and the motion branch has 114M parameters.
Increasing the number of CMA layers and placing them later in the network generally improves performance, and A is the best among tested settings. See \cref{sec:appendix:mot-arch} for further analysis.
}
    \vspace{-2pt}
\resizebox{0.74\linewidth}{!}{
\begin{tabular}[t]{@{}lccccccc @{}}

\toprule
 & R@1 $\uparrow$ & R@2$\uparrow$ & R@3$\uparrow$ 
 & FID$\downarrow$ & MMDist$\downarrow$ & DIV$\rightarrow$ & MModality$\uparrow$
 \\ 
\midrule
Real & 0.518 & 0.713 & 0.813 & - & 2.811 & 9.976 & - \\
\midrule
$A$ & \textbf{0.536} & \textbf{0.728} & \textbf{0.819} & 0.241 & \textbf{2.767} & 10.379 & 2.454 \\ 
$B_{\cn 1}$ & 0.502 & 0.707 & 0.807 & 0.311 & 2.895 & 10.318 & 2.489 \\ 
$B_{\cn 2}$ & 0.508 & 0.714 & 0.812 & 0.288 & 2.8637 & 10.261 & 2.315 \\ 
$B_{\cn 3}$ & 0.508 & 0.712 & 0.811 & 0.22 & 2.879 & 10.405 & 2.664 \\ 
$B_{\cn 4}$ & \underline{0.514} & \underline{0.716} & \underline{0.814} & 0.243 & \underline{2.839}& 10.347 & 2.534 \\ 
$C_{\cn 1}$ & 0.506 & 0.701 & 0.801 & 0.236 & 2.894 & 10.386 & 2.684 \\ 
$C_{\cn 2}$ & 0.502 & 0.702 & 0.795 & 0.285 & 2.948 & 10.333 & 2.631 \\ 
$C_{\cn 3}$ & 0.503 & 0.705 & 0.803 & \underline{0.171} & 2.886 & 10.221 & 2.819 \\ 
$D_{\cn 1}$ & 0.473 & 0.663 & 0.767 & 0.283 & 3.105 & \textbf{10.176} & \textbf{3.770} \\ 
$D_{\cn 2}$ & 0.477 & 0.672 & 0.777 & \textbf{0.164} & 3.092 & \underline{10.189} & \underline{3.197} \\ 
\bottomrule
\end{tabular}
}
\label{tab:appendix:mot-attn}
  \vspace{-8pt}
\end{table}

\newpage
\subsection{Ablation on Model Size}
\label{sec:appendix:abl-size}
  \vspace{-4pt}

We examine model size along three axes while keeping all other settings fixed: (i) the overall capacity, achieved by scaling the text and motion branches proportionally, (ii) motion-branch capacity with a fixed text branch, and (iii) language-backbone size with a comparable motion branch.

\cref{tab:abl-modelsize-t2m} compares overall \textbf{backbone sizes}. With roughly 3$\times$ parameters, the medium model yields modest gains on R@k and MMDist despite slightly higher FID.
This suggests greater capacity helps capture high-level semantics, though realizing its full benefit may require careful optimization.

\begin{figure}[t]
    \centering
    \includegraphics[width=\linewidth]{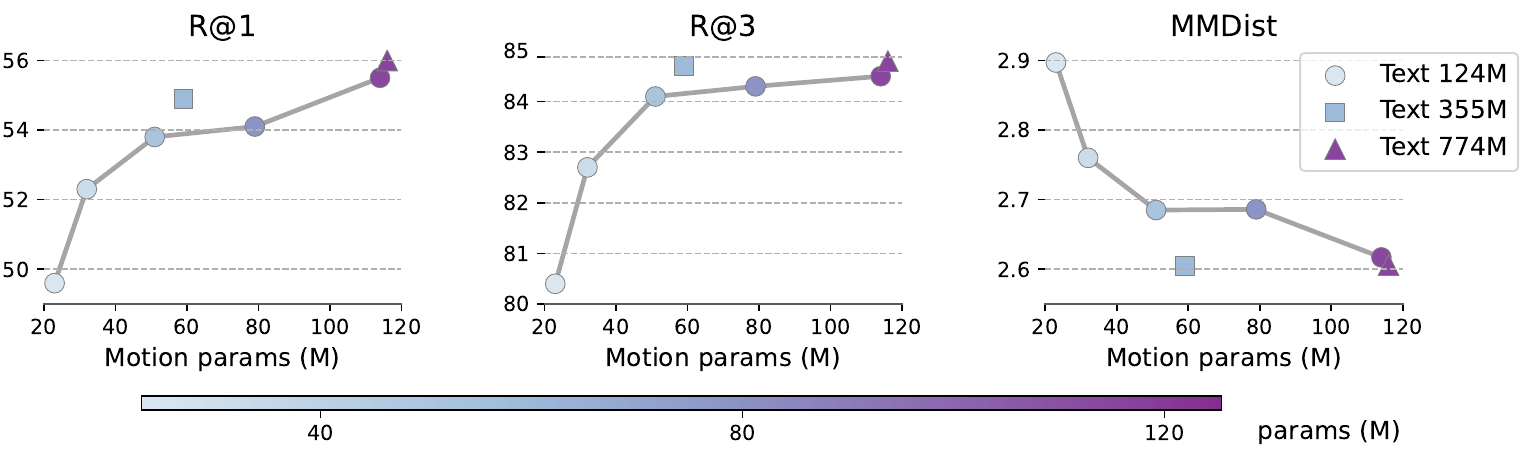}
    \vspace{-6pt}
    \caption{
    Ablation on branch capacity on motion generation.
    All models are trained for 200K iterations.
    A 124M text branch already performs competitively with much larger backbones (355M/774M).
    Our model can achieve competitive performance with only halfed motion parameters ($\sim$51M).
    }
    \vspace{-12pt}
    \label{fig:appendix:mot-param}
\end{figure}

With the text branch fixed to GPT-2 small (124M), we \textbf{scale the motion branch} from 23M to 114M parameters by setting hidden size to {76, 192, 384, 576, 768} (\cref{fig:appendix:mot-param}).
Increasing motion capacity generally improves text–motion alignment (higher R@k) and reduces MMDist, while diversity remaining stable. in a high level
A half-sized motion branch ($\sim$51M) already offers a strong trade-off, delivering competitive overall performance, and, the best FID among the 124M-text configurations.

\textbf{Text-branch Size.}
To isolate the effect of the language backbone, we replace GPT-2 small (124M) with GPT-2 medium (355M) and GPT-2 large (774M), keeping the motion branch of comparable size (pairs: 355M/59M vs. 124M/51M, and 774M/116M vs. 124M/114M).
Larger text branches further improve alignment and lower MMDist, and tend to increase diversity/MModality.

\begin{table}[h]
    \centering
    \caption{
    Effect of GPT-2 backbone size and CFG guidance scale $\omega$ on text-to-motion task. All models are trained for 200K iterations. 
    The medium backbone, with more parameters (692M v.s. 238M) consistently outperforms the small model.
    }
    \label{tab:abl-modelsize-t2m}
 \vspace{-2pt}
\resizebox{\columnwidth}{!}{
    \begin{tabular}{@{}l ccc cccccc@{}}
    \toprule
        
        Size & Text Params & Total Params & $\omega$
        & R@1 $\uparrow$& R@2 $\uparrow$& R@3 $\uparrow$& FID $\downarrow$& MMDist$\downarrow$ & DIV$\rightarrow$ 
        \\
    \midrule
        small & 124M & 238M & 1.0
         & $0.534$ & $0.739$ & $0.842$ & $0.222$ & $2.61$ & $10.256$
         \\
        medium & 355M  & 692M & 1.0
         & $0.558$ & $0.756$ & $0.852$ & $0.235$ & $2.553$ & $10.238$
         \\
    \midrule
        small & 124M & 238M  & 3.0
         & $0.552$ & $0.759$ & $0.852$ & $0.173$ & $2.554$ & $10.239$
         \\
        medium & 355M & 692M  & 3.0
         & $0.568$ & $0.766$ & $0.860$ & $0.192$ & $2.489$ & $10.084$
         \\
    \bottomrule
    \end{tabular}
}
\end{table}

\vspace{-4pt}
\subsection{VAE and Diffusion Head}
\label{sec:appendix:vae-abl}

\textbf{Ablation on Diffusion Head.}
We ablate the diffusion head $\mathcal{H}$ in our motion branch to study how 
conditioning design affects generation.
We vary (i) supervision with a diffusion head $\mathcal{H}$ v.s. direct MSE regression,
(ii) the mapping from backbone hidden states to the diffusion condition (multi-head attention, MHA vs. linear layer),
(iii) the number of motion holders $h\_num\in\{1,4,8\}$ used to query hidden states from the autoregressive backbone, 
and (iv) classifier-free guidance (CFG) at sampling.
All variants are trained for 200K iterations with results in \cref{tab:appendix:abl-vae}.

We observe that 
(i) Direct MSE supervision (the last row) yields the weakest performance, confirming the benefit of diffusion-based training.
(ii) Increasing $h\_num$ (b-d) enriches the conditioning signal and improves retrieval accuracy R@k, but also raises FID, suggesting a harder denoising problem. 
A moderate setting of $h\_num =4$ offers the best trade-off. 
(iii) With $h\_num=4$ and no CFG (c)(e), an MHA head outperforms a linear mapper, achieving higher R@3 (0.529 v.s. 0.521), lower MMDist ( 2.645 v.s.2.713), and lower FID (0.166 v.s. 0.178)
The advantage is even larger under sparse conditioning ($h\_num=1$ in (d)(e), FID: 0.164 vs.\ 0.283);
(iv) Enabling CFG on the same configuration further improves both alignment and fidelity. 
Accordingly, we adopt diffusion supervision with an MHA head and $h\_num=4$ as the default.

\textbf{Guidance scale.} 
We sweep the CFG guidance scale $\omega$ on the unified model (results in \cref{tab:appendix-tmr}). 
Note that guidance is applied only to \emph{motion generation}. 
Moderate guidance performs best: $\omega=5.0$ minimizes FID on the T2M, while $\omega=3.0$ yeilds the best R-Precision and MultiModal Distance.
Very weak ($\omega=1$) or overly strong ($\omega\geq10$) guidance degrades alignment and diversity. We thus use $\omega=4$ in all main results.

\begin{table}[h]
  \centering
  \caption{
  Ablation of motion generation head and loss on HumanML3D. All variants are trained on T2M task for 200k iterations, with same backbone and data.
  We vary: (i) supervision with diffusion head $\mathcal{H}$ (Diff.) or direct MSE regression (MSE), (ii) mapping of multi-head attention (MHA) or Linear layer in $\mathcal{H}$, (iii) number of motion holder <motion\_out> ($h\_num$) (i.e., the count of hidden states passed from the backbone to $\mathcal{H}$), and (iv) classifier-free guidance (CFG) at sampling.
  }
  \scriptsize
  \resizebox{.98\columnwidth}{!}{%
\begin{threeparttable}
\begin{tabular}{@{}l l ccc ccc cccc@{}}
  \toprule
  ID & Loss & $h\_num$ & Head & CFG 
  & R@1 $\uparrow$ & R@2 $\uparrow$ & R@3 $\uparrow$ 
  & FID $\downarrow$ & MMDist $\downarrow$ & DIV $\rightarrow$ & MModality $\uparrow$ \\
  \midrule
    \rowcolor{gray!20}
    (a)  & 
    Diff. &
    MHA & 4 & \Checkmark 
    & \textbf{0.547} & \textbf{0.751} & \textbf{0.850} & \textbf{0.149} & \textbf{2.578} & 10.041 & {2.265} 
    \\
    (b) & 
    Diff. &
    MHA & 8 & \XSolidBrush 
    & \underline{0.531} & \underline{0.733} & 0.836 & 0.185 & 2.655 & 10.154 & 2.198
    \\
    (c) & 
    Diff. &
    MHA & 4 & \XSolidBrush 
    & 0.529 & 0.730 & \underline{0.839} & 0.166 & \underline{2.645} & 10.012 & 2.350
    \\
    (d) & 
    Diff. &
    MHA & 1 & \XSolidBrush 
    & 0.525 & 0.729 & 0.831 & \underline{0.164} & 2.678 & 10.090 & 2.514 
    \\
    (e) & 
    Diff. &
    Linear & 4 & \XSolidBrush 
    & 0.521 & 0.731 & 0.829 & 0.178 & 2.713 & 9.985 & \underline{2.603} 
    \\
    (f) & 
    Diff. &
    Linear & 1 & \XSolidBrush 
    & 0.525 & 0.729 & 0.829 & 0.283 & 2.689 & 10.069 & \textbf{2.719} 
    \\
    (g) & 
    MSE &
    Linear & 4 & - 
    & 0.518 & 0.725 & 0.823 & 0.276 & 2.705 & 9.758 & 2.175 
    \\
    
 \bottomrule
 \end{tabular}
  {
     \footnotesize
    Following MotionGPT’s evaluation protocol~\citep{jiang2023motiongpt}, results are averaged over two runs.
    
    Best results are \textbf{bold}, second-best are \underline{underlined}. 
    The default configuration is {\sethlcolor{gray!20} \hl{gray-shaded}}.
  }
\end{threeparttable}
}
  \label{tab:appendix:abl-vae}
 \vspace{-8pt}
\end{table}

\begin{table}[h]
 \centering
 \scriptsize
 \caption{Ablation on guidance scale $\omega$ in CFG, for motion latent diffusion on HumanML3D, with model trained on unified tasks. 
 Best and second-best results are \textbf{bold} and \underline{underlined}.
 }
 \label{tab:abl-cfg}
 \vspace{-4pt}
\resizebox{0.72\columnwidth}{!}{%
 \begin{threeparttable}
 \begin{tabular}{@{}lccccccc@{}}
 \toprule
    $\omega$ & R@1 & R@2 & R@3 & FID$\downarrow$ & MMDist $\downarrow$ & DIV$\rightarrow$ & MModality$\uparrow$ \\
    \midrule
    Real & $0.519$ & $0.724$ & $0.820$ & 0.002 & $2.753$ & $9.941$ & -  \\
    \midrule
    $1$ & $0.534$ & $0.727$ & $0.828$ & $0.143$ & $2.714$ & $10.086$ & $\textbf{1.717}$ \\
    $2$ & \textbf{0.555} & \underline{$0.754$} & $0.843$ & $0.123$ & $2.601$ & $10.006$ & $1.321$ \\
    $3$ & $\underline{0.554}$ & $\textbf{0.756}$ & $\textbf{0.850}$ & $0.103$ & $\textbf{2.585}$ & $\textbf{9.926}$ & $1.272$ \\
    \rowcolor{gray!20}
    $4$ & $0.552$ & $0.753$ & \underline{$0.848$} & \underline{$0.098$} & \underline{$2.589$} & \underline{$9.911$} & $1.258$ \\
    $5$ & $0.552$ & $0.752$ & $0.848$ & $\textbf{0.094}$ & $2.593$ & $9.906$ & $1.248$ \\
    $6$ & $0.546$& $0.751$ & $0.849$ & $\textbf{0.094}$ & $2.598$ & $9.900$ & $1.243$ \\
    $10$ & $0.546$ & $0.748$ & $0.844$ & $0.109$ & $2.620$ & $9.870$ & $1.281$ \\
    $15$ & $0.541$ & $0.739$ & $0.839$ & $0.12$ & $2.653$ & $9.873$ & $1.312$ \\
    $20$ & $0.533$ & $0.728$ & $0.826$ & $0.134$ & $2.739$ & $9.827$ & \underline{$1.385$} \\
   
    \bottomrule
  \end{tabular}
 {
     \footnotesize
    Following MotionGPT’s protocol, results are averaged over two runs.
    
    The default configuration is {\sethlcolor{gray!20} \hl{gray-shaded}}.
    }
\end{threeparttable}
 }
 \vspace{-8pt}
\end{table}

\subsection{Effectiveness of Training Scheme}
\label{sec:appendix:stage}

We adopt a three-stage schedule (see \cref{sec:train}, \cref{fig:training}): SI, text-to-motion (T2M) pretraining; SII, cross-modal alignment with joint optimization on T2M and motion-to-text (M2T) (SII); and SIII, joint fine-tuning. 
We evaluate three settings: (i) \textbf{Three Stages}, the full three-stage schedule, (ii) \textbf{Two Stage}, a two-stage schedule without SI, and (iii) \textbf{Trained Text Branch}, a two-stage variant in which the text branch is unfrozen during SI–SII, rendering SIII unnecessary.
We report results on both generation (T2M) and understanding (M2T) in \cref{tab:appendix:training}.

\begin{table}[t]
\scriptsize
\centering
\caption{Training-scheme evaluation on HumanML3D~\citep{Guo_2022_CVPR_humanml3d}, with protocol in ~\citet{jiang2023motiongpt}. 
Stage1: T2M Pre-training, Stage2: Cross-Modal Alignment, Stage3: Joint Fintuning.
\Checkmark marks enabled stages, while colors encode the state of text branch, {\color{textC}updated} or {frozen}.
Jointly updating the text branch from the start improves early T2M in SI but degrades final T2M after SII and markedly lowers M2T scores (rows “Trained Text Branch”). 
}
\resizebox{.96\columnwidth}{!}{%
\begin{threeparttable}
    \begin{tabular}{@{}lccc ccccccc@{}}
  \toprule
    \multirow{2}{*}{Type} & 
    \multirow{2}{*}{Stage1} & 
    \multirow{2}{*}{Stage2} & 
    \multirow{2}{*}{Stage3} & 
    \multicolumn{3}{c}{Text-to-Motion}& 
    \multicolumn{3}{c}{Motion-to-Text} 
    \\ 
    \cmidrule(lr){5-7} \cmidrule(lr){8-10} 
    &
    &&& R@3 $\uparrow$ & FID $\downarrow$ & MMDist $\downarrow$ & R@1$\uparrow$ & Bleu@4$\uparrow$ & BertScore$\uparrow$ 
    \\ 
  \midrule
    \multirow{3}{*}{Three Stages} &
    \Checkmark & \XSolidBrush & \XSolidBrush
    & 0.826 & 0.239 & 2.797
    & - & - & -
    \\
    & 
    \Checkmark &  \Checkmark  & \XSolidBrush
    & \underline{0.831} & \underline{0.215} & \underline{2.755}
    & 0.571 & 18.328 & 33.993
    \\
    & 
    \Checkmark &  \Checkmark  & {\color{textC}\Checkmark}
    & \textbf{0.837} & \textbf{0.208} & \textbf{2.725}
    & 0.573 & 19.412 & 35.231
    \\
  \midrule
    \multirow{2}{*}{Two Stages} &
    \XSolidBrush & \Checkmark & \XSolidBrush
    & 0.755 & 0.298 & 3.213
    & 0.561 & 18.295 & 34.676
    \\
    & 
    \XSolidBrush & \Checkmark & {\color{textC}\Checkmark}
    & 0.772 & 0.325 & 3.108
    & 0.573 & 18.277 & 35.546
    \\
  \midrule
    \multirow{2}{*}{Trained Text Branch} &
    {\color{textC}\Checkmark} & \XSolidBrush & -  
    & 0.822 & 0.239 & 2.832
    & - & - & -
    \\
    & 
    {\color{textC}\Checkmark}  & {\color{textC}\Checkmark}  & - 
    & 0.801 & 0.243 & 2.942
    & 0.505 & 14.119 & 33.385
    \\ 
  
  \bottomrule
    \end{tabular}%
    Following MotionGPT’s protocol, results are averaged over two runs. \\
    Best and second-best results are \textbf{bold} and \underline{underlined}.
\end{threeparttable}
}
\label{tab:appendix:training}
\end{table}

\textbf{Text-to-motion Pre-training} and \textbf{Cross-Modal Alignment}.  
Pretraining on T2M (SI) yields strong motion generation and provides a motion-specialized initialization.
Entering SII confers M2T capability and further improves T2M (alignment improves and MMDist/FID drop), indicating that explicit cross-modal alignment benefits both directions.
Training directly with multi-task objectives from scratch (i.e., without SI) markedly degrades T2M quality, even after subsequent joint optimization, underscoring the importance of a motion-specific warm start (i.e., initialization from T2M pretraining).

\textbf{Joint Fine-tuning.}
Once SII has established cross-modal alignment, SIII yields modest gains, primarily stabilizing M2T while preserving T2M, and thus serves as a light refinement.
The “Two Stages” variants (SII+SIII without SI) show that joint optimization can boost both tasks when the model is under-initialized.
However, it also degrades the language branch’s competence, leading to worse T2M and M2T than the full three-stage schedule.
Although incremental gains beyond SI+SII are modest, SIII can mitigate residual negative transfer and calibrate cross-modal alignment under noisy or shifted training conditions, yielding more stable results.

\textbf{Freezing v.s. training the text branch.} 
To promote modality-specific representations and reduce negative transfer onto a well-trained language branch, we propose to freeze the text branch in SI–SII.
Freezing preserves linguistic competence while the motion branch specializes.
By contrast, updating all parameters from the start (“Trained Text Branch”) can give slightly higher T2M scores in SI (e.g., R@3 0.834 vs. 0.820; MMDist 2.698 vs. 2.787), but after SII these models exhibit degraded T2M and notably weaker M2T (e.g., BertScore 0.713; Bleu@4 3.577), consistent with 'catastrophic forgetting'.
We attribute this decline to negative transfer from the new motion branch onto the text branch during early training. In practice, keeping the LM frozen lets the motion branch learn a more stable, modality-specific space and achieve more reliable alignment under \textit{limited} paired data.

{

}

In summary, SI provides essential motion-specific initialization; SII delivers the bulk of cross-modal gains; SIII offers small, stabilizing improvements. Freezing the text branch through SI–SII prevents loss of linguistic ability and yields the best overall balance between understanding and generation.


\section{Details on MotionGPT3}
\label{sec:appendix:detail}

\begin{figure}[ht] 
    \centering
    \includegraphics[width=.95\linewidth]{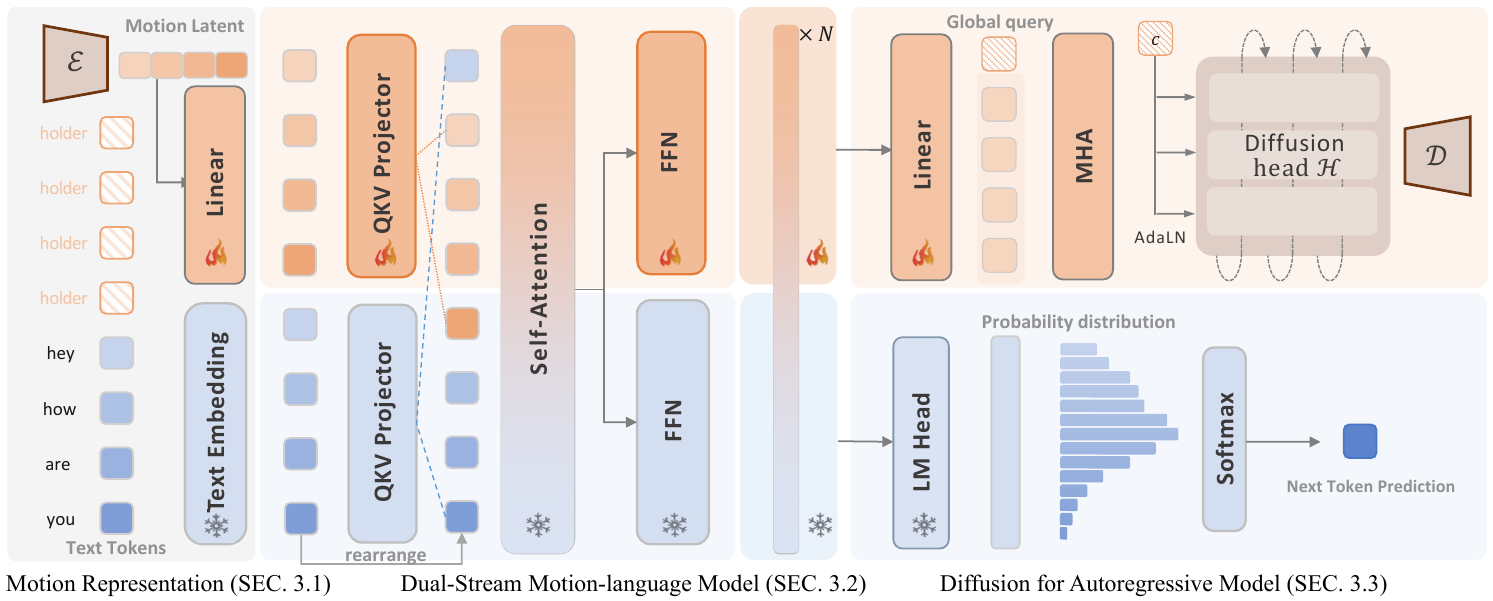}
    \vspace{-1pt}
    \caption{Details of our bimodal motion-language model. 
    {\color{motionC} Motion} and {\color{textC} text} inputs are encoded by separate branches and then reordered to their original sequence order before cross-modal self-attention.
    After $N$ hybrid layers, text is generated autoregressively by next-token prediction, while motion is produced via a diffusion head $\mathcal{H}$.
    Panels correspond to motion representation (\cref{sec:vae}), the dual-stream motion–language backbone (\cref{sec:mot}), and the diffusion module (\cref{sec:diff}).
    }
    \label{fig:arch}
\end{figure}

\subsection{Motion Generation in Unified Model}  
\label{sec:appendix:detail-generation}
\textbf{Diffusion Loss} 
We use a diffusion head $\mathcal{H}$ to map backbone hidden states into a denoised motion latent. As illustrated in \cref{fig:training,fig:arch}, we insert $K$ \textit{motion holder} tokens {\color{motionC} <motion\_out>} as queries to extract the corresponding output states from the motion branch. 
Let the backbone hidden size be $d_t$ and the motion latent size be $d_m$ (we set $d_t{=}768$, $d_m{=}256$ in our default model). 
After one forward pass, the queried states form $h \in \mathbb{R}^{K \times d_t}$.
An MHA pooling module aggregates $h$ and produces a global condition vector $c \in \mathbb{R}^{1 \times d_m}$
via an internal mapping to match the motion-latent dimensionality,
which is then fused with the timestep embedding $\tau(t)$ to condition $c$.
With the cumulative product of the noise schedule $\bar{\alpha}_t \!=\! \prod_{s=1}^{t} \alpha_s$,  
, we sample $t \sim \mathcal{U}\{1,\dots,T\}$ and corrupt the ground-truth motion latent $z_0 \in \mathbb{R}^{d_m}$ by
\begin{equation}
z_t=\sqrt{\bar{\alpha}_t}\, z_0+\sqrt{1-\bar{\alpha}_t}\,\epsilon,\quad
\epsilon\sim\mathcal{N}(0,I),
\end{equation}
Given a noisy latent $z_t \in \mathbb{R}^{1 \times d_m}$ at timestep $t$, the head predicts the noise $\hat{\epsilon}_\theta $
and is trained with the standard $\epsilon$-prediction objective
\begin{equation}
    L_{diff}\;=\;\mathcal{E}_{z_0,\epsilon,t}\lVert\epsilon - \hat{\epsilon}_\theta(\alpha_tz_0+\sigma_t\epsilon,t,c)\rVert_2^2,\quad
    \hat{\epsilon}_\theta \!=\! \mathcal{H}(z_t,t,c)
\end{equation}
where $\alpha_t$, $\sigma_t$ follow a linear schedule equivalent to the forward noising process.
At inference, we start from $z_T \!\sim\! \mathcal{N}(0,I)$ iteratively denoise with the sampler of $\mathcal{H}$ down to $t{=}0$ to obtain $\hat{z}_0$, which is then decoded by the motion decoder $\mathcal{D}$ into a raw motion sequence.


\textbf{Architecture of $\mathcal{H}$.}
The diffusion head first processes the $K$ queried hidden states with a TransformerEncoderLayer and aggregates them via multi-head attention pooling into a single condition vector $c$.
We fuse $c$ with the timestep embedding and modulate each block via AdaLN.
$\mathcal{H}$ consists of a stack of 1024-wide residual blocks.
Each block applies AdaLN followed by a two-layer MLP with SiLU nonlinearity, and the final block projects to the noise prediction $\hat{\epsilon}_\theta$.


\textbf{Cross-Entropy Loss for Boundary Tokens}
To delimit motion from text decoding, we introduce two boundary tokens{\color{motionC} <som>} (start of motion) and {\color{motionC}<eom>} (end of motion). 
At inference, once the LM predicts {\color{motionC}<som>} via next-token prediction, we generate the motion latent in a single forward pass, with $K$ {\color{motionC} <motion\_out>} holders concatenated to the sequence, and then append {\color{motionC} <eom>} deterministically.
During training, we apply cross-entropy only to the {\color{motionC}<som>} prediction, and {\color{motionC} <eom>} is not supervised.

\newpage
\subsection{Motion VAE}
\label{sec:appendix:detail-vae}

\textbf{AutoEncoder}
We adopt a Transformer-based motion VAE~\citep{chen2023mld} with an encoder $\mathcal{E}$ and a decoder $\mathcal{D}$ that maps an $M$-frame motion sequence $m^{1:M}$ to a compact continuous latent $z \in \mathbb{R}^{n\times d}$ ($n=1, d=256$) and reconstructs the motion via $m^{1:M}=\mathcal{D}(z)=\mathcal{D}(\mathcal{E}(m^{1:M})$.
Both $\mathcal{E}$ and $\mathcal{D}$ are transformers~\citep{vaswani2017attention} with long skip connection~\citep{ronneberger2015u}, and without the action biases used in~\citet{petrovich21actor}.
This design yields an expressive latent space that supports accurate semantic understanding and high-fidelity, diverse motion synthesis.



\textbf{Architecture}
Given an input motion sequence $m^{1:M}$ of length $M$, the encoder $\mathcal{E}$ processes the sequence together with a small set of learnable distribution tokens and outputs the Gaussian parameters $(\mu_m,\sigma_m)$.
A latent $z$ is sampled by reparameterization $z=\mu_m+\sigma_m\epsilon, \epsilon in \mathcal{N}(0,I)$. 
The decoder $\mathcal{D}_{mld}$ performs cross-attention over the latent vector $z$ to query $L$ motion tokens, which are then projected back into $\hat{m}^{1:L}$ in the raw motion space.
\begin{equation}
    \hat{m}^{1:L} = \mathcal{D}(\mathcal{E}(m^{1:M})
\end{equation}


\textbf{Loss}
We train the motion VAE with reconstruction term over framewise poses and KL regularizer on the latent, following standard practice in VAEs~\citep{kullback1951information,kingma2013auto}.
Let $m^{1:M}$ denote a sequence of $M$ ground-truth poses $m_t \in \mathbb{R}^{263}$ and $\hat m^{1:M}$ the decoder outputs.
The encoder produces a Gaussian posterior $q_\phi(z\,|\,m^{1:M})=\mathcal{N}(\mu,\mathrm{diag}(\sigma^2))$.
The objective is
\begin{equation}
\label{eq:vae_total_loss}
\mathcal{L} \;=\; \mathcal{L}_{\mathrm{rec}} \;+\; \lambda_{\mathrm{KL}}\, \mathcal{L}_{\mathrm{KL}},
\end{equation}
with
\begin{equation}
\label{eq:vae_rec}
\mathcal{L}_{\mathrm{rec}}
\;=\; \frac{1}{T}\sum_{t=1}^{T} \bigl\| m_t - \hat m_t \bigr\|_2^2,
\end{equation}
\begin{equation}
\label{eq:vae_kl}
\mathcal{L}_{\mathrm{KL}}
\;=\; D_{\mathrm{KL}}\!\left(\mathcal{N}(\mu,\mathrm{diag}(\sigma^2)) \,\big\|\, \mathcal{N}(0,I)\right).
\end{equation}
For completeness, the KL term admits the closed form
\(
\mathcal{L}_{\mathrm{KL}}
=\tfrac{1}{2}\sum_{j}\!\left(\mu_j^2 + \sigma_j^2 - \log \sigma_j^2 - 1\right).
\)


\textbf{Raw Motion Representation}
Following ~\citet{Guo_2022_CVPR_humanml3d}, each frame $m^i\in\mathbb{R}^{263}$ concatenates
a tuple of root angular velocity $\dot{r}^a\in\mathbb{R}$ along Y-axis, root linear velocities $(\dot{r}^x,\dot{r}^z)\in\mathbb{R}$ on XZ-plane, root height $r^y\in\mathbb{R}$, local joints positions $j^p\in\mathbb{R}^{3N_j}$, velocities $j^v\in\mathbb{R}^{3N_j}$, and rotations $j^r\in\mathbb{R}^{6N_j}$ in root space, with $N_j$ denotes the joint number, and binary foot ground contact features $c^f\in\mathbb{R}^{4}$ by thresholding the heel and toe joint velocities. This finally results in $m^i = \{\dot{r}^a, \dot{r}^x, \dot{r}^z, r^y, j^p, j^v, j^r, c^f\}$.

\subsection{Metrics Definitions}
\label{sec:appendix:metrics}
We adopt standard metrics for text–motion alignment, motion quality and diversity, caption quality, and (for VAE analysis) motion reconstruction. Unless noted, features are computed with the official HumanML3D/T2M evaluator ~\citep{guo2022generating}, with motion encoder $\phi(m)$ and text encoder $\psi(t)$.

\paragraph{Text–motion alignment}
To evaluate semantic consistency between generated motions and input texts, we adopt motion-text retrieval precision (R-Precision) at Top-k(R@k), and the Multimodal Distance (MMDist), which measures the embedding-space distance between paired modalities.
\textbf{R@k} measures retrieval accuracy within a candidate set: for each query (text or motion), we rank candidates of the other modality by cosine similarity and report the fraction of cases where the paired item appears in the top-k.
\textbf{MMDist} is the average embedding distance between paired items:
\begin{equation}
 \text{MMDist} = \frac{1}{N} \Sigma_{n=1}^{N}\lVert\phi(m_n) - \psi(t_n)\rVert_2
\end{equation}

\paragraph{Motion quality}
\textbf{FID} assess how closely generated motions match ground truth ones in feature space, indicating overall quality, and is computed between the Gaussian fits of $\{\phi(m)\}$ for generated and ground-truth motions in the evaluator feature space.

\paragraph{Diversity}
Diversity (DIV) measures feature variation across samples, and MultiModality (MM), which quantifies variation among motion generations from the same textual description. 
Following \citet{guo2022generating,chen2023mld}, we randomly sample all generated motions into two subsets, $\{x_i\}_{i=0}^{X_d}$ and $\{x_i'\}_{i=0}^{X_d}$, of the same size $X_d$. 
Then \textbf{DIV} is formalized as:
\begin{equation}
    \text{DIV} = \frac{1}{X_d}\Sigma_{i=1}^{X_d} \lVert x_i-x_i' \rVert
\end{equation}
Randomly sample a set of text descriptions with size $J_m$ and sample two subsets of $X_m$ motions generated by $j$-th for each text description, denote as $\{x_{j,i}\}_{i=0}^{X_m}$ and $\{x_{j,i}'\}_{i=0}^{X_m}$. \textbf{MM} is calculated as:
\begin{equation}
    \text{MM} = \frac{1}{J_m \times X_m}\Sigma_{i=1}^{J_m}\Sigma_{i=1}^{X_m} \lVert x_{j,i}-x_{j,i}' \rVert
\end{equation}

\paragraph{Motion captioning}
We follow prior work ~\t{chuan2022tm2t} and adopt standard NLP metrics including BLEU~\citep{papineni2002bleu}, ROUGE-L~\citep{lin2004rouge}, CIDEr~\citep{vedantam2015cider}, and BERTScore~\citep{zhang2019bertscore} to evaluate the fluency, relevance, and diversity of generated captions.

\paragraph{Reconstruction}
We evaluate reconstruction fidelity of motion autoencoders with:
\textbf{MPJPE} and \textbf{PAMPJPE} for global and local errors in millimeter,
\textbf{ACCL} (Acceleration Error) computed from second-order finite differences,
and \textbf{APE/AVE} (Absolute Position/Velocity Error) reported over {root}, {trajectory}, {pose}, and {joints} components.

%% file: sections/LLM.tex
 \section{The Use of Large Language Models (LLMs)}
We used ChatGPT only for grammar/typo checks; all technical content, experiments, and analyses were authored and verified by the authors without substantive LLM contribution.